\documentclass[10pt,twocolumn,letterpaper]{article}

\usepackage{iccv}
\usepackage{times}
\usepackage{epsfig}
\usepackage{graphicx}
\usepackage{amsmath}
\usepackage{amssymb}

\usepackage[utf8]{inputenc}
\usepackage{times}
\usepackage{graphicx}
\usepackage{amsfonts, amsthm}
\usepackage{color}
\usepackage{caption}
\usepackage{subcaption}
\usepackage{xspace}
\usepackage{paralist}
\usepackage{import}
\usepackage{array,booktabs,calc}
\usepackage{placeins}
\usepackage{cite}

\usepackage[pagebackref=true,breaklinks=true,letterpaper=true,colorlinks,bookmarks=false]{hyperref} 

\newcommand{\bc}{\mathbf{c}}
\newcommand{\bD}{\mathbf{D}}

\newcommand{\bI}{\mathbf{I}}

\newcommand{\bK}{\mathbf{K}}

\newcommand{\bp}{\mathbf{p}}

\newcommand{\bR}{\mathbf{R}}
\newcommand{\bs}{\mathbf{s}}
\newcommand{\bt}{\mathbf{t}}
\newcommand{\bu}{\mathbf{u}}

\newcommand{\bX}{\mathbf{X}}

\newcommand{\bz}{\mathbf{z}}

\newcommand{\bmu}{\boldsymbol{\mu}}

\newcommand{\bsigma}{\boldsymbol{\sigma}}

\newcommand{\nR}{\mathbb{R}}

\newcommand{\cL}{\mathcal{L}}

\newcommand{\cN}{\mathcal{N}}

\newcommand{\cS}{\mathcal{S}}

\newcommand{\cZ}{\mathcal{Z}}

\newcommand{\figref}[1]{Fig.~\ref{#1}}
\newcommand{\secref}[1]{Section~\ref{#1}}

\newcommand{\tabref}[1]{Table~\ref{#1}}

\makeatletter
\DeclareRobustCommand\onedot{\futurelet\@let@token\@onedot}
\def\@onedot{\ifx\@let@token.\else.\null\fi\xspace}
\def\eg{e.g\onedot} 
\def\ie{i.e\onedot} 
 
\def\etc{etc\onedot}

\def\wrt{wrt\onedot}

\def\etal{et~al\onedot}

\makeatother

\newcommand{\boldparagraph}[1]{\vspace{0.2cm}\noindent{\bf #1:} }

\definecolor{darkgreen}{rgb}{0,0.7,0}

\iccvfinalcopy % *** Uncomment this line for the final submission

\begin{document}

%%%%%%%%% TITLE
\title{Texture Fields: Learning Texture Representations in Function Space}

\author{Michael Oechsle$^{1, 2}$ \quad Lars Mescheder$^{1}$ \quad Michael Niemeyer$^{1}$ \quad Thilo Strauss$^{2 \dagger}$ \quad Andreas Geiger$^1$\\
$^1$Autonomous Vision Group, MPI for Intelligent Systems and University of Tübingen \\
$^2$ETAS GmbH, Bosch Group, Stuttgart \\
{\tt\small$^{1}$\{firstname.lastname\}@tue.mpg.de} \qquad 
{\tt\small$^{\dagger}$\{firstname.lastname\}@etas.com}}
\maketitle

%%%%%%%%% ABSTRACT

\begin{abstract}
In recent years, substantial progress has been achieved in learning-based reconstruction of 3D objects. At the same time, generative models were proposed that can generate highly realistic images. However, despite this success in these closely related tasks, texture reconstruction of 3D objects has received little attention from the research community and state-of-the-art methods are either limited to comparably low resolution or constrained experimental setups. A major reason for these limitations is that common representations of texture are inefficient or hard to interface for modern deep learning techniques. In this paper, we propose Texture Fields, a novel texture representation which is based on regressing a continuous 3D function parameterized with a neural network. Our approach  circumvents limiting factors like shape discretization and parameterization, as the proposed texture representation is independent of the shape representation of the 3D object. We show that Texture Fields are able to represent high frequency texture and naturally blend with modern deep learning techniques. Experimentally, we find that Texture Fields compare favorably to state-of-the-art methods for conditional texture reconstruction of 3D objects and enable learning of probabilistic generative models for texturing unseen 3D models. We believe that Texture Fields will become an important building block for the next generation of generative 3D models.

\end{abstract}

%%%%%%%%% BODY TEXT
\begin{figure}[t]
	\centering
	\includegraphics[width=8cm]{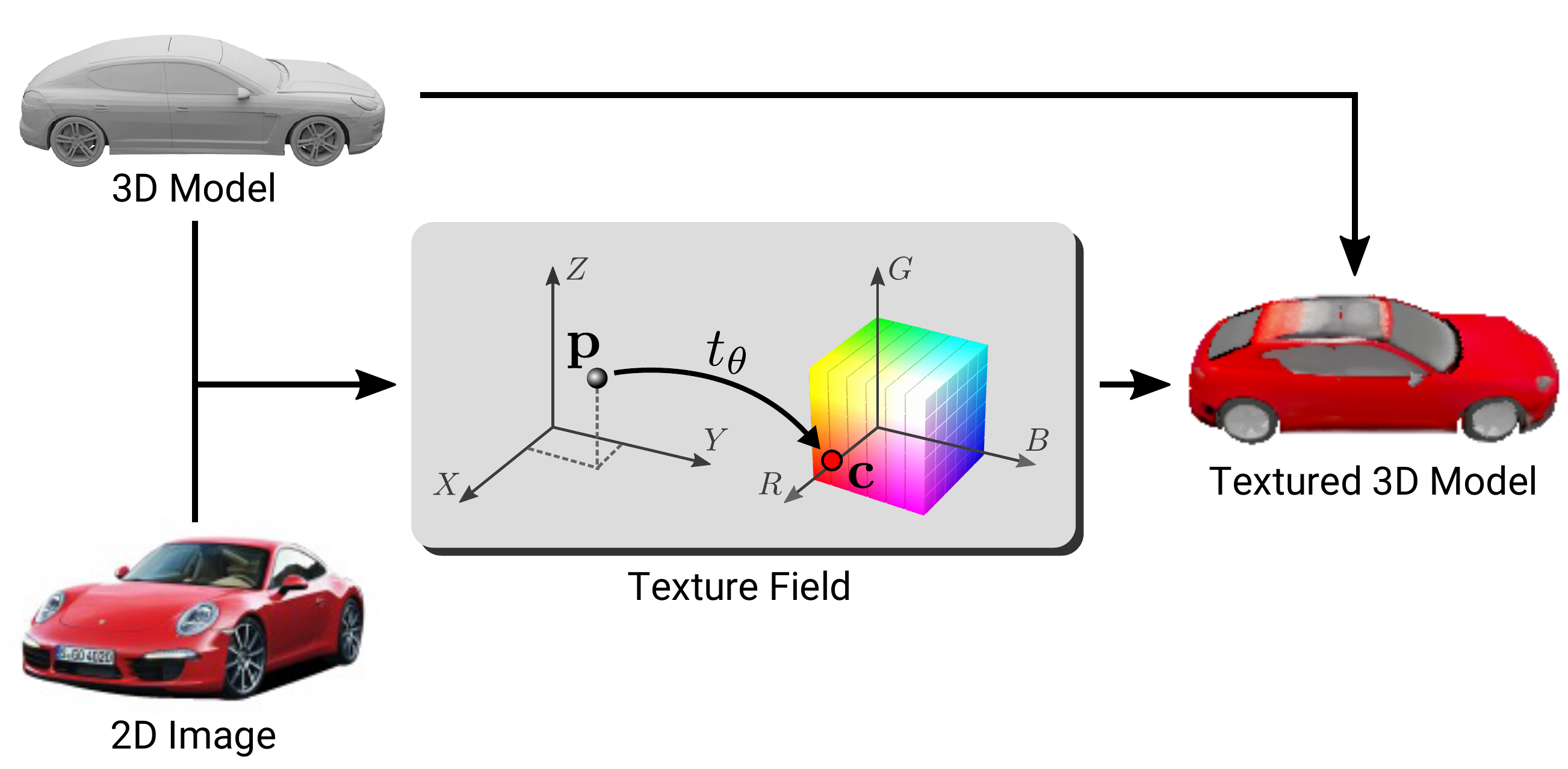}
	\caption{\textbf{Texture Fields} take a 3D shape and (optionally) a single 2D image of an object as input and learn a continuous function $t_\theta$ which maps any 3D point $\bp$ to a color value $\bc$, enabling the prediction of accurately textured 3D models.
	}
	\label{fig:teaser}
	\vspace{-0.5cm}
\end{figure}

\section{Introduction}

3D reconstruction is one of the grand goals of computer vision. 
Recently, the vision community has witnessed impressive progress in reconstruction tasks like single image 3D reconstruction \cite{Choy2016ECCV,Wang2018ECCV,Groueix2018CVPR,Mescheder2019CVPR,Chen2019CVPR} and generating 3D objects \cite{Wu2016NIPS,Rezende2016NIPS,Brock2016ARXIV} using learning-based techniques which resolve ambiguities by incorporating prior knowledge.
However, previous work on learning-based 3D reconstruction has mainly focused on the problem of reconstructing geometry.
In contrast, texture reconstruction of 3D objects has received less attention from the research community.

Previous approaches for texture reconstruction from single images differ in how texture information is represented. 
Several recent works \cite{Tulsiani2017CVPR,Sun2018ARXIV} have proposed volumetric voxel representations for colored 3D reconstruction. 
Unfortunately, however, due to the computational cost of voxel representations, which grows cubically with the resolution, the voxelizations are limited to rather low resolutions (usually $32^3$ or $64^3$)
and hence cannot represent high frequency details.
An alternative representation of texture consists of a 2D texture atlas and a parameterized mesh using a UV-mapping that maps a point on the shape manifold to a pixel in the texture atlas. 
However, current approaches based on mesh representations usually assume known topology and a predefined template mesh which limits these approaches to specific object categories such as birds \cite{Kanazawa2018ECCV} or faces \cite{Saito2017CVPR}.
Predicting texture in the general case for arbitrary shapes without template mesh remains an unsolved problem. 

\boldparagraph{Contributions}
The main limiting factor of the aforementioned methods is their texture representation which either does not allow for high resolution reconstruction or strongly relies on task-specific shape parameterizations, limiting generality of these methods. 
An ideal texture representation, in contrast, should be independent of the shape representations (voxels, point cloud, mesh, \etc.) and able to represent high frequency detail.  
Towards this goal, we propose Texture Fields as a novel representation of texture. 
Our key idea is to learn a continuous function for representing texture information in 3D space, see \figref{fig:teaser}.
By parameterizing this function through a deep neural network we are able to integrate this representation into a deep learning pipeline for 3D texture reconstruction that can be trained end-to-end. 

Our experiments on various 3D object categories demonstrate
that Texture Fields are able to predict high frequency texture information from a single image.
We combine our approach with a state-of-the-art 3D reconstruction method \cite{Mescheder2019CVPR}, yielding a method which jointly reconstructs both the 3D shape and the texture of an object.
Besides conditional experiments, we also extend our novel texture representation to the generative setting and show preliminary results for texture synthesis given a 3D shape model and a latent texture code. 
We conduct experiments on texture transfer between different objects as well as interpolations in the latent texture space which demonstrate that our generative model learns a useful representation of texture.	  
	
\section{Related Work}
\label{Sec:RelWork}

We now briefly discuss the most related works on single image 3D reconstruction, generative image modeling using 3D knowledge and continuous representation learning.

\subsection{Single Image Reconstruction}

In recent years, 3D reconstruction of shapes from single images has made great progress \cite{Choy2016ECCV,Wang2018ECCV,Groueix2018CVPR,Mescheder2019CVPR,Chen2019CVPR}. 

\boldparagraph{Voxels}
Tulsiani \etal \cite{Tulsiani2017CVPR} proposed a voxel-based texture representations for learning colored 3D reconstruction based on ray consistency and multi-view supervision. 
More recently, Sun \etal \cite{Sun2018ARXIV} combined 2D-to-3D appearance flow estimation and voxel color regression for learning to reconstruct colored
voxelizations in a supervised fashion.
Unfortunately, however, voxel-based representations are severely limited in resolution due to computational and memory constraints.
In contrast, our continuous approach does not require discretization and thus results in more details compared to voxel-based representations as we show in our experimental evaluation in \secref{sec:results}.

\boldparagraph{Point Clouds}
In recent years, novel view synthesis \cite{Zhou2016ECCV, Rematas2017TPAMI} has been used for reconstructing colored point clouds from single images \cite{Tatarchenko2016ECCV,Park2017CVPR}.
By combining novel view synthesis and depth estimation, the method proposed in \cite{Tatarchenko2016ECCV} reprojects the predicted image into a colored point cloud.
In \cite{Park2017CVPR}, point clouds are reconstructed from a set of novel views using multi-view stereo algorithms.
Unfortunately, point-cloud based representations are sparse.
While dense mesh representation can be extracted from point clouds, the reconstructed shapes typically do not match the quality of state-of-the-art 3D reconstruction approaches and inferring the missing texture information requires additional post-processing steps.
In contrary, our approach allows for inferring appearance for any location in 3D space and can be used in combination with arbitrary shape representations.

\boldparagraph{Meshes}
Mesh-based approaches rely on category-specific template models and rigid texture parameterizations \cite{Kanazawa2018ECCV, Saito2017CVPR}. 		
In contrast, our reconstruction approach can represent texture for arbitrary meshes without requiring a UV texture map of a category-specific template model.
Note that determining a proper UV-mapping for arbitrary mesh models is a non-trivial problem which is typically solved with various heuristics for atlas generation.
The advantage of our approach is that we circumvent mesh parameterization by disentangling the texture from the shape representation.

\subsection{Generative Models}

Recent work has shown that image generation methods \cite{Goodfellow2014NIPS,Radford2016ICLR,Karras2018ICLR,Mescheder2018ICML,Brock2019ICLR,Karras2018ARXIV} can be improved by exploiting 3D knowledge about the generated shapes \cite{Zhu2018NIPS,Alhaija2018ACCV,Alhaija2018IJCV}. 
Alhaija \etal \cite{Alhaija2018IJCV} propose a model which learns to translate intrinsic properties (depth, normals, material) into RGB images.
Zhu \etal \cite{Zhu2018NIPS} learn to predict 3D geometry as well as texture information in 2D images by disentangling shape, texture and pose in an unsupervised setting. 
In contrast to those image-based approaches, we directly predict the texture for the entire object in 3D.

\subsection{Continuous Representations}

Recently, parameterized continuous functions gained popularity for 3D shape reconstruction.
Several works \cite{Mescheder2019CVPR,Chen2019CVPR,Park2019CVPR} proposed to formulate 3D reconstruction as learning a continuous occupancy function or a signed-distance field in 3D space, parameterized via a neural network.
Furthermore, in \cite{Garnelo2018ICMLWORK}, a continuous function for color values was used for generating 2D images.
In this work, we extend the concept of learning a representation in function space to reconstruct and generate texture information of 3D objects.

%\clearpage

\begin{figure*}[t!]
	\includegraphics[width=\linewidth]{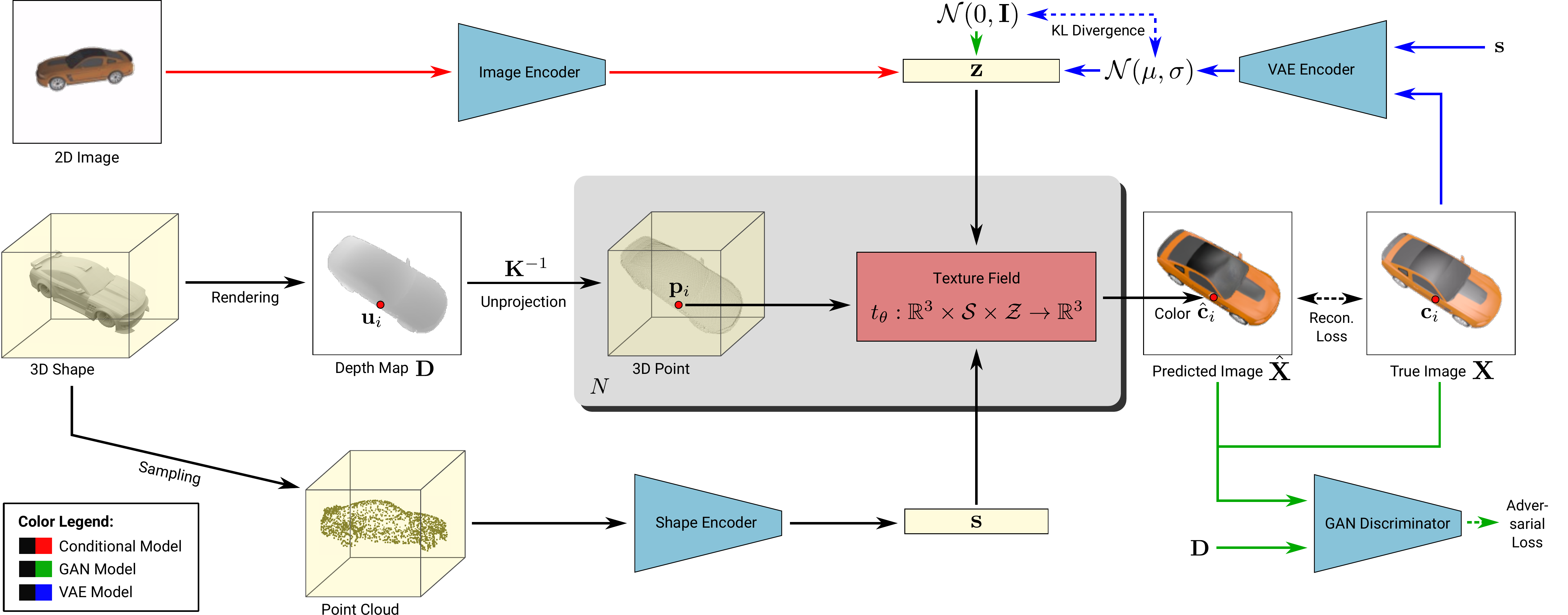}
	\caption{\textbf{Model Overview.} Colored arrows show \textit{alternative} pathways representing the \textcolor{red}{Conditional}, \textcolor{darkgreen}{GAN} and \textcolor{blue}{VAE} model. The blue and red boxes denote trainable components of our model which are parameterized through neural networks, respectively.
	The 3D shape of the object is encoded into a fixed-length vector representation $\bs$. Additionally, we render a depth map $\bD$ from a randomly chosen viewpoint and obtain the corresponding 3D points $\bp_{i}$ by unprojecting all $N$ image pixels $\bu_{i}$ into 3D. The reconstruction loss minimizes the difference between the pixel colors predicted by the Texture Field $\hat{\bc}_{i}=t_\theta(\bp_{i},\bs,\bz)$ and the ground truth pixel colors $\bc_{i}$. For training a \textcolor{red}{Conditional} model, the latent variable $\bz$
	encodes information about the input image. In the unconditional case (\ie, for the \textcolor{darkgreen}{GAN} and \textcolor{blue}{VAE}), $\bz$ is sampled from a Gaussian distribution.}
	\label{fig:model_overview}
	\vspace{-0.3cm}
\end{figure*}

\section{Method}

This section describes the proposed Texture Field representation and demonstrates how it can be applied to conditional and unconditional texture synthesis tasks.

\subsection{Texture Fields}

Recent approaches to 3D reconstruction~\cite{Mescheder2019CVPR,Chen2019CVPR,Park2019CVPR} represent 3D shapes as continuous functions of occupancy probability or surface distance.
In contrast to point-, voxel- or mesh-based representations, these approaches do not rely on a fixed discretization and thus form an ideal basis for representing appearance information.
We explore this idea by embedding surface texture as a continuous function in 3D space.
In combination with state-of-the-art 3D reconstruction techniques this allows us to reconstruct a textured 3D model from a single 2D image.

Let $t(\cdot)$ denote a function which maps a 3D point $\bp \in \mathbb{R}^3$ to a color value $\bc \in \mathbb{R}^3$ hence representing texture information by means of a 3D vector field:
\begin{equation}
t:\nR^3 \rightarrow \nR^3
\end{equation} 
By parameterizing the function $t$ as a neural network $t_\theta$ with learnable parameters $\theta$, we reduce the problem to a simple regression task.
However, the problem of texture generation remains ill-posed without any further constraints.
Thus, in order to inform $t_\theta$ about the input shape of the object, we condition $t_\theta$ on a shape embedding $\bs\in\cS$. This helps the Texture Field to guide its predictions towards the object surface and allows for exploiting contextual geometric information such as surface discontinuities which are often aligned with image edges.
As input shape, we explore 3D CAD models as well as image-based 3D reconstructions using neural networks \cite{Mescheder2019CVPR} in this paper.

Unfortunately, even given the input 3D shape, there still exists a variety of plausible texture explanations.
Consider cars, for instance, where the 3D geometry alone does neither determine the color, nor the exact shape of the windows or headlights.
However, we may further constrain the task by providing information about the object appearance.

More specifically, we condition $t_\theta$ on an additional 2D image taken from an arbitrary viewpoint.
Note that an image provides only partial appearance information as it only depicts the object from a single perspective.
Furthermore, we encode the image into a viewpoint-invariant global feature representation $\bz\in\cZ$. Thus, our approach does not assume the camera extrinsics \wrt the object to be known and therefore is able to texture untextured 3D shapes using images ``in the wild''.
Importantly, the input image need not depict an object of the exact same shape as the 3D model. This would be a strong restriction in practice as often only an approximate shape can be retrieved from a single image.

In summary, we define a \emph{Texture Field} as a mapping from 3D point $\bp$, shape embedding $\bs$ and condition $\bz$ to color $\bc$:
\begin{equation}
t_\theta: \nR^3 \times \cS \times \cZ \rightarrow \nR^3
\end{equation}
In the following, we will consider both the conditional case as well as the unconditional case.
For the unconditional case, we exploit probabilistic generative networks \cite{Kingma2014ICLR,Goodfellow2014NIPS}, capturing ambiguity in a \textit{random} latent code $\bz$.

\boldparagraph{Model}
An overview over our \textit{Texture Field} model is shown in \figref{fig:model_overview}.
Colored arrows indicate alternative pathways representing the \textcolor{red}{Conditional}, \textcolor{darkgreen}{GAN} and \textcolor{blue}{VAE} model.
Blue and red boxes denote the trainable components of our model which are parameterized by neural networks whose parameters are trained jointly.
We now provide details on each of the components of our model.

\boldparagraph{Shape Encoder}
To inform the Texture Field $t_\theta$ about the 3D shape of the object, we uniformly sample 3D points from the input shape (typically provided in form of a traingular mesh) and pass the resulting point cloud to a PointNet encoder \cite{Qi2017CVPR}, adopting the network architecture of \cite{Mescheder2019CVPR}. This results in a fixed-dimensional shape embedding $\bs$.

\boldparagraph{Image Encoder}
For the conditional case (\textcolor{red}{red} arrows in \figref{fig:model_overview}), we additionally extract appearance information from an input image. More specifically, we encode the input image into a fixed-dimensional latent code $\bz$ using a standard pre-trained residual network \cite{He2016CVPR} with 18 layers.

\boldparagraph{Texture Field}
Given the input shape $\bs$ and the condition $\bz$, our proposed Texture Field model is able to predict a color value $\bc_i$ for any 3D point $\bp_i$. Thus, it would be possible to directly color every point on the 3D mesh. Unfortunately, mesh-based appearance representations require additional UV-mappings for representing texture. We therefore train our texture model in 2D image space which provides a regular and hence efficient representation\footnote{Note that this does not restrict our model in any way. At test time, our model can be evaluated for arbitrary 3D points.}.

Towards this goal, we render depth maps $\bD$ and corresponding color images $\bX$ from arbitrary viewpoints using OpenGL. The color at pixel $\bu_i$ and depth $d_i$ is predicted as
\begin{equation}
\hat{\bc}_i = t_\theta \left(d_i\,\bR\,\bK^{-1} \bu_i +\bt,\bs,\bz\right)
\end{equation}
where $i$ denotes the index for pixels with finite depth values $i\in\{1, \dots, N\}$ and $N$ refers to the number of foreground pixels in the image (\ie, pixels where the object is visible).
Here, the camera intrinsics and extrinsics are denoted by $\bK\in\nR^{3\times 3}$ and ($\bR\in\nR^{3\times 3},\bt\in\nR^3$), respectively, and pixel $\bu_i$ is represented in homogeneous coordinates.
For training our model, we compare the predicted color $\hat{\bc}_i$ to the color of the corresponding pixel $\bc_i$ in the rendered image $\bX$.

\subsection{Training}
\label{sec:training}

This sections describes how we train our conditional and unconditional models. See \figref{fig:model_overview} for a visual illustration.

\boldparagraph{Conditional Setting}
In the conditional case, we input an embedding $\bz$ of the input image to our network\footnote{It is important to note that the input image and the images $\bX$ used for supervising the model during training need \textit{not} be the same as we learn a viewpoint-invariant representation $\bz$.}. 
We train $t_\theta (\bp,\bs,\bz)$ in a supervised setting by minimizing the $\ell_1$-loss between the predicted image $\hat{\bX}$ and the rendered image $\bX$:
\begin{equation}
\cL_{cond}= \frac{1}{B} \sum_{b=1}^{B} \sum^{N_b}_{i=1} {\Vert t_\theta (\bp_{bi}, \bs_b, \bz_b) - \bc_{bi}\Vert}_{1}
\label{eq:conditional_loss}
\end{equation}
Here, $B$ represents the mini-batch size.
Each element of the mini batch represents an image with $N_b$ foreground pixels (\ie, pixels where the object is visible). Note that the shape encoding $\bs_b$ and the input encoding $\bz_b$ implicitly depend on the parameters of the shape and image encoder networks, respectively. We train the parameters of all three networks (shape encoder, image encoder, Texture Field) jointly.

\boldparagraph{Unconditional Generative Model} 
\label{sec:unconditionalsetting}
In the unconditional setting we are only given the 3D shape as input but no additional information about the appearance of the object.
As mentioned above, this is a highly ill-posed task with many \textit{valid} explanations.
We therefore utilize probabilistic generative models which capture the ambiguity in the output using a latent code $\bz$ sampled from a Gaussian distribution (see \textcolor{darkgreen}{green} and \textcolor{blue}{blue} models in \figref{fig:model_overview}).
In particular, we adapt two recent deep latent variable models to our setting: a generative adversarial networks (GAN) \cite{Goodfellow2014NIPS} and a variational auto-encoder (VAE) \cite{Kingma2014ICLR}.	
Both models have been applied to a variety of image-based tasks in the past 
\cite{Goodfellow2014NIPS,Radford2016ICLR,Karras2018ICLR,Kingma2014ICLR,Rezende2014ICML,Gregor2015ICML}
and are thus ideally suited in the context of our image-based loss formulation.

Let us first consider adversarial training.
The problem of learning a generative model for texture information given a 3D shape can be tackled using a \emph{conditional GAN} \cite{Mirza2014ARXIV} where the generator is conditioned on the 3D shape.
The generator is represented as a Texture Field $t_\theta: \nR^3 \times \cS \times \cZ \rightarrow \nR^3$ which maps the latent code $\bz$ for every given 3D location $\bp_i$ conditioned on the shape embedding $\bs$ to an RGB image:
\begin{equation}
\hat{\bX} = G_\theta(\bz_b|\bD_b,\bs_b) = \{t_{\theta}(\bp_{bi},\bs_b,\bz_b)|i\in\{1,\dots,N_b\}\}
\end{equation}
As above, $b$ denotes one element of the mini-batch and $\bs_b$ depends on the parameters of the shape encoder.
We use a standard image-based discriminator $D_\phi (\bX_b|\bD_b)$ conditioned on the input depth image $\bD$ by concatenating it with the input image.
For training the model, we use a non-saturating GAN loss with $R_1$-regularization \cite{Mescheder2018ICML}.  

An alternative method for learning a latent variable model is given by a \emph{conditional VAE} (cVAE) \cite{Sohn2015NIPS}.
Our cVAE model comprises an encoder network that maps color image $\bX$ to mean $\bmu$ and variance $\bsigma$ of an isotropic normal distributed random variable $\bz$ which follows the distribution $q_\phi(\bz|\bX,\bs)$. 
Our Texture Field model is used as decoder by predicting the color value $\hat{\bc}_i$ of each pixel $\bu_i$ for the corresponding 3D locations $\bp_i$ conditioned on the shape embedding $\bs$.
Following \cite{Kingma2014ICLR,Rezende2014ICML}, we minimize the variational lower bound 
\begin{align}
\cL_{\text{VAE}} = \frac{1}{B} \sum_{b=1}^{B} \left[ \vphantom{\sum^{N_b}_{i=1}} \beta \; \text{KL}(q_\phi(\bz|\bX_b,\bs_b) \parallel p_0(\bz_b)) \right. \nonumber\\
\left. + \sum^{N_b}_{i=1} {\Vert t_\theta(\bp_{bi}, \bs_b, \bz_b)- \bc_{bi}\Vert}_{1} \right]
\label{eq:vae-objective}
\end{align}
where KL refers to the Kullback-Leibler divergence, $p_0(\bz) = \cN(\bz|0,\bI)$ denotes the standard normal distribution,
$\bz_b$ is a sample from the posterior distribution $q_\phi(\bz|\bX_b,\bs_b) $
and $\beta$ is a trade-off parameter between the KL-divergence and the reconstruction loss \cite{Higgins2017ICML}.
Again, $b$ denotes one element of the mini-batch and $\bs_b$ depends on the parameters of the shape distribution.
In practice, we set $\beta = 1$.
During training, we optimize $\cL_{\text{VAE}}$ using the reparameterization trick~\cite{Kingma2014ICLR,Rezende2014ICML}.
At test time, we obtain new texture samples for the 3D object under consideration by sampling $\bz$ from a standard normal distribution.

\subsection{Implementation details}
In all of our experiments, we implement the texture field $t_\theta(\cdot, \bs, \bz)$ using the fully connected ResNet~\cite{He2016CVPR} architecture from \cite{Mescheder2019CVPR}, see supplementary for details.
For the image encoder, we use a ResNet-18 architecture~\cite{He2016CVPR}, pretrained on ImageNet.
For the shape encoder, we adopt the PointNet \cite{Qi2017CVPR} architecture from \cite{Mescheder2019CVPR}.
The GAN discriminator and the VAE encoder are based on the discriminator from \cite{Mescheder2018ICML}.
We train both our supervised model and our VAE model end-to-end using Adam \cite{Kingma2015ICLR} with learning rate $10^{-4}$. 
Our GAN is trained with alternating gradient descent using the RMSProp Optimizer \cite{Tieleman2012} with the same learning rate.

\section{Experimental Evaluation}
\label{sec:results}

We evaluate our approach in three different experiments.
In the first part, we investigate the \textbf{representation power} of Texture Fields by analyzing how well the method can represent high frequency textures when trained on a single 3D object.
In the second part, we apply our method to the challenging task
of \textbf{single view texture reconstruction} in which we
predict the full texture of 3D objects given only the 3D shape and a single view of this object.
Moreover, we combine our method with a state-of-the-art shape reconstruction method \cite{Mescheder2019CVPR} for full textured 3D reconstruction which also includes reconstructing the shape of the 3D object.
In the last experiment, we explore if our representation can also be used in a \textbf{generative setting} where the goal is to produce a varied distribution of possible textures given only the shape of the 3D object without further input.

\boldparagraph{Baselines}
Only few prior works have considered texture reconstruction
in the general case without predefined template mesh or information about the camera view.
Therefore, we construct a first simple baseline by mapping the input image onto the mesh by projecting all vertices into the input view to determine their color. 
While this simple baseline receives additional information about the camera of the input view, we believe it still serves as a good sanity check to see if our method is actually learning something useful and if it is able to correctly fill-in occluded regions.
As a second baseline we consider a novel-view-synthesis (NVS) approach which uses the same image encoder as our approach, but applies a UNet architecture\footnote{%
See supplementary material for details.}
\cite{Ronneberger2015MICCAI} to transform a depth rendering of the object into an RGB image.
While this approach can also generate novel views of the object under consideration, it requires additional (lossy) post-processing to generate a complete texture map of the object.
In particular, there is no guarantee for this baseline that the newly generated views are consistent under viewpoint changes.
Lastly, we consider Im2Avatar~\cite{Sun2018ARXIV} as a baseline for full textured 3D reconstruction from single images.
To the best of our knowlege, this method currently achieves the highest resolution ($64^3)$ among all voxel-based 3D reconstruction methods which predict colors and produces state-of-the-art results. We use the official implementation\footnote{%
\url{https://github.com/syb7573330/im2avatar}
} of Im2Avatar.

\boldparagraph{Dataset}
Unless specified otherwise, we use the categories `cars', `chairs', `airplanes' and `tables' from the ShapeNet dataset~\cite{Chang2015ARXIV} as these categories contain rich texture information while also providing a large variety of shapes.
For our conditional experiments, we use the renderings provided by Choy~\etal~\cite{Choy2016ECCV} as input.
For training our models, we render 10 additional images and depth maps per object from random views in the upper hemisphere.

\boldparagraph{Metrics}
For evaluation, we consider three different metrics in image space for random views of the objects.
Firstly, to evaluate how well our method and the baselines capture the appearance distribution for a given object category, we use the Fréchet Inception Distance (FID)~\cite{Heusel2017NIPS}. This is a common metric between distributions of images and widely used in the GAN literature~\cite{Heusel2017NIPS, Karras2018ICLR, Karras2018ARXIV, Brock2019ICLR}.
Moreover, to better estimate the distance between a predicted view and the ground truth on a per-instance basis, we measure the structure similarity image metric (SSIM)~\cite{Wang2004TIP}.
Since we found that SSIM mostly captures local properties of images, we additionally introduce a more global perceptual measure, which we call \emph{Feature-$\ell_1$-metric}.
Similarly to FID, the Feature-$\ell_1$-metric is computed by embedding both the image produced by the method under consideration and the ground truth view into feature space using an Inception network.
The Feature-$\ell_1$-metric then computes the mean absolute distance between the feature activations of the predicted and the ground truth image.

\subsection{Representation Power}

\begin{figure}[t!]
\centering
\begin{tabular}{c@{}c@{}c}
	\includegraphics[width=0.3\linewidth,trim={5cm 0.5cm 5cm .5cm},clip]{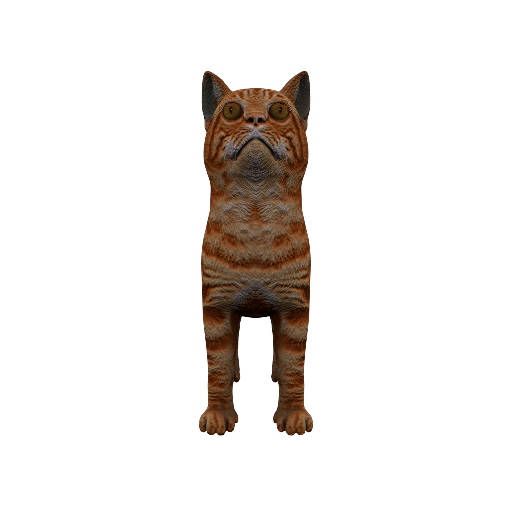} &
	\includegraphics[width=0.3\linewidth,trim={5cm 0.5cm 5cm .5cm},clip]{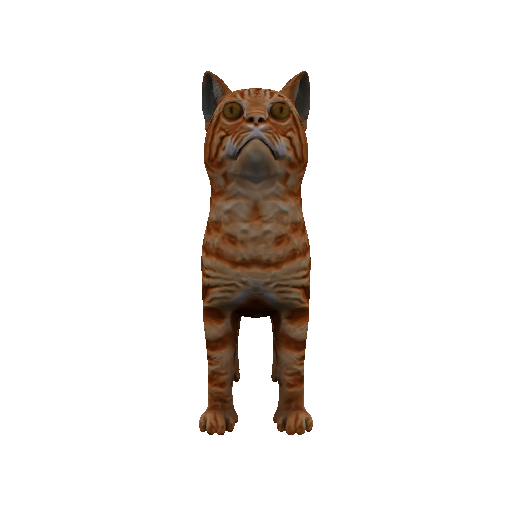} & 
	\includegraphics[width=0.3\linewidth,trim={10cm 1cm 10cm 1cm},clip]{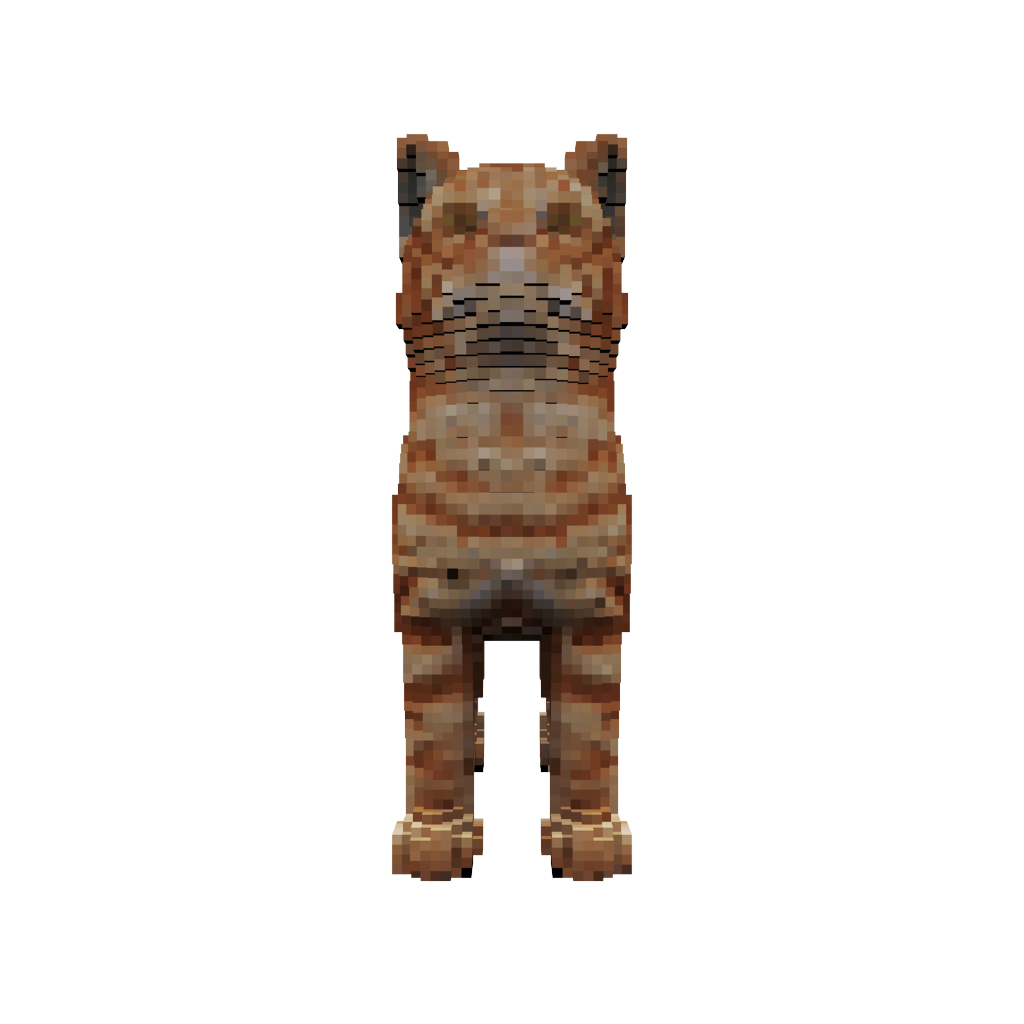} \\
	\includegraphics[width=0.3\linewidth,trim={5cm 0.5cm 5cm 0.2cm},clip]{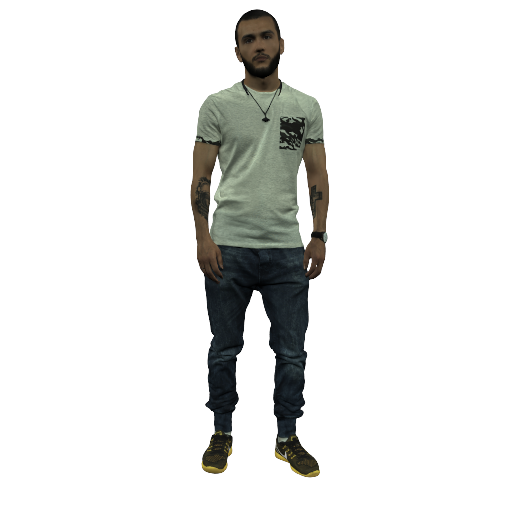} &
	\includegraphics[width=0.3\linewidth,trim={5cm 0.5cm 5cm 0.2cm},clip]{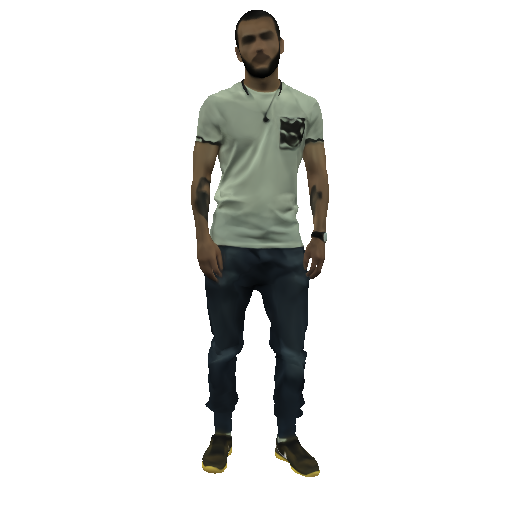} &
	\includegraphics[width=0.3\linewidth,trim={10cm 1cm 10.cm 0.4cm},clip]{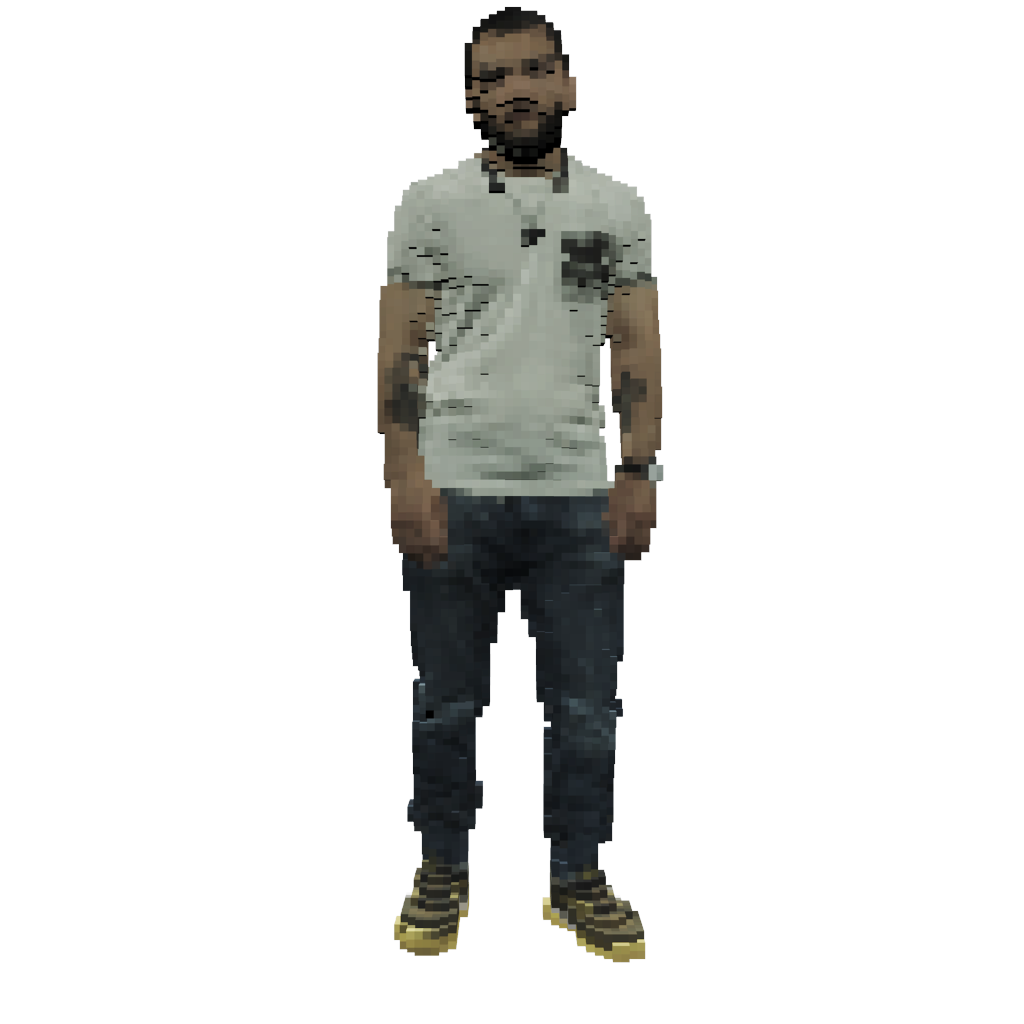} \\
	Ground Truth &  Texture Field  &  Voxelization ($128^3$) 
\end{tabular}
\caption{
\textbf{Representation Power.}
Comparison of a Texture Field fitted to a 3D model of a cat/human to a voxel-based representation and the corresponding ground truth model.
}
\label{fig:representation_power}
\end{figure}

The goal of this experiment is to explore the representation power of our Texture Field
model. 
This ``overfitting'' experiment helps to disentangle the quality of the image encoder from the Texture Field representation itself and provides an upper bound on the reconstruction quality that we can expect when applying Texture Fields to more difficult tasks.
We train our method separately on 3D meshes of a cat and a human\footnote{3D models from \url{free3d.com}
and \url{turbosquid.com}.},
wrt. $512^2$ px renderings 
from 500 views.

Qualitative results are shown in \figref{fig:representation_power}, comparing our model to a voxelization at a fixed resolution of $128^3$ voxels.
We observe that Texture Fields can represent high-frequency information while the voxel representation is inherently restricted to a Manhattan world of limited granularity.
This experiment validates that Texture Fields are a promising texture representation.

\subsection{Single Image Texture Reconstruction}
\label{subsec:experiments.singleimage}

We now turn our attention to the challenging task of single-image texture reconstruction.
Towards this goal, we conduct experiments for texture reconstruction given a 3D model together with a 2D image of the object from a random camera view.
We train our method using the conditional setup described in \secref{sec:training}.
At test time we apply the trained model in three different settings.
In the first setting, we use ground truth 3D shapes as input and apply our model to synthetic renderings of the object. 
In the second setting, we combine our method with a state-of-the art shape reconstruction approach \cite{Mescheder2019CVPR} to obtain a full  3D reconstruction pipeline of both shape and textures.
Finally, in the third setting, we investigate if our model also transfers to real data.
Here, we use images of real cars together with similar 3D shapes from the ShapeNet dataset.

\begin{figure}[t!]
\centering
\begin{tabular}{c|c@{}c@{}c}
    & Projection & NVS & Texture Field  \\
   \includegraphics[width=0.22\linewidth,trim={0.5cm 1.5cm 0.5cm 2.5cm},clip]{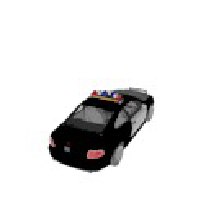} &
   \includegraphics[width=0.22\linewidth,trim={0.5cm 1.5cm 0.5cm 2.5cm},clip]{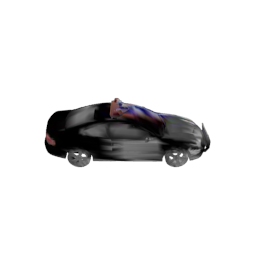} & 
   \includegraphics[width=0.22\linewidth,trim={0.5cm 1.5cm 0.5cm 2.5cm},clip]{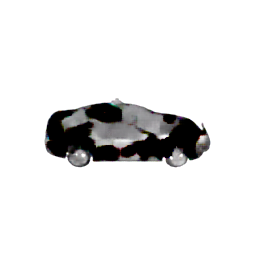} &
   \includegraphics[width=0.22\linewidth,trim={0.5cm 1.5cm 0.5cm 2.5cm},clip]{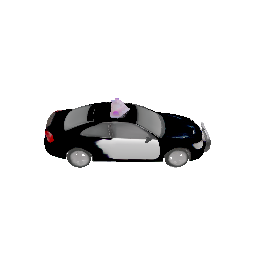} \\
   \includegraphics[width=0.22\linewidth,trim={0.5cm .cm 0.5cm 0.5cm},clip]{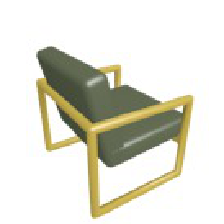} &
   \includegraphics[width=0.22\linewidth,trim={0.5cm .5cm 0.5cm 0.5cm},clip]{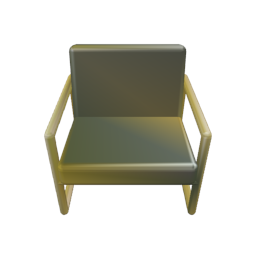} & 
   \includegraphics[width=0.22\linewidth,trim={0.5cm .5cm 0.5cm 0.5cm},clip]{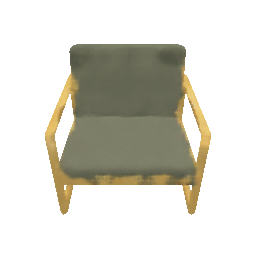} &
   \includegraphics[width=0.22\linewidth,trim={0.5cm .5cm 0.5cm 0.5cm},clip]{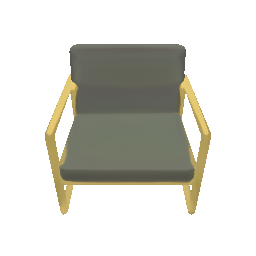} \\
   \includegraphics[width=0.22\linewidth,trim={0.7cm 1.5cm 0.5cm 2.5cm},clip]{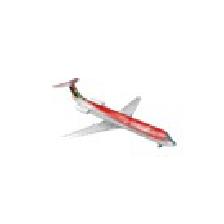} &
   \includegraphics[width=0.22\linewidth,trim={0.7cm 1.5cm 0.5cm 2.5cm},clip]{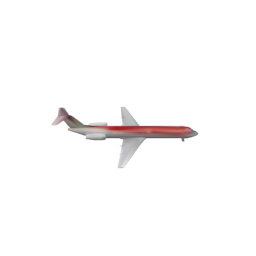} & 
   \includegraphics[width=0.22\linewidth,trim={0.7cm 1.5cm 0.5cm 2.5cm},clip]{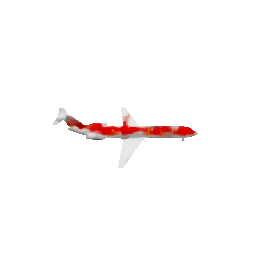} &
   \includegraphics[width=0.22\linewidth,trim={0.7cm 1.5cm 0.5cm 2.5cm},clip]{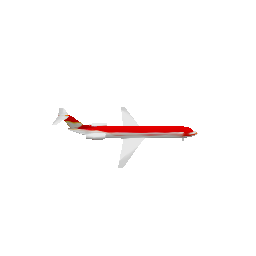} \\
   \includegraphics[width=0.22\linewidth,trim={0.5cm 0.5cm 0.5cm 1.5cm},clip]{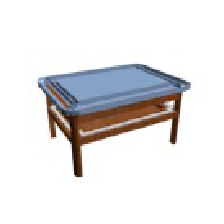} &
   \includegraphics[width=0.22\linewidth,trim={0.5cm 0.5cm 0.5cm 1.5cm},clip]{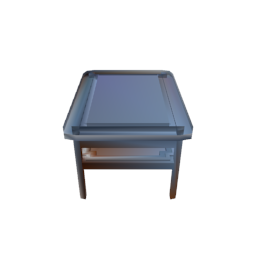} & 
   \includegraphics[width=0.22\linewidth,trim={0.5cm 0.5cm 0.5cm 1.5cm},clip]{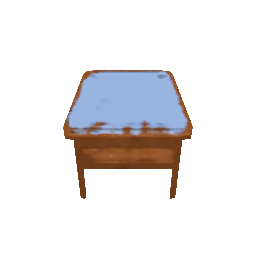} &
   \includegraphics[width=0.22\linewidth,trim={0.5cm 0.5cm 0.5cm 1.5cm},clip]{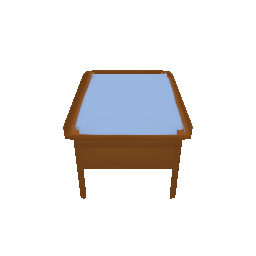} \\
    \;Condition (2D) & \multicolumn{3}{c}{Prediction (3D)}  \;
   \end{tabular}
\caption{
\textbf{Texture Reconstruction.} A qualitative comparison between our method (Texture Field) and the baselines is shown. In this experiment, we use GT shapes as input.  
}
\label{fig:single_image_gt}
\vspace{-0.3cm}
\end{figure}
	
\begin{table*}[t!]
	\centering
	\resizebox{\textwidth}{!}{
		\begin{tabular}{l|ccc|ccc|ccc}
\toprule
{} & \multicolumn{3}{c}{FID} & \multicolumn{3}{c}{SSIM} & \multicolumn{3}{c}{Feature-$\ell_1$} \\
{} & Projection &     NVS &    Texture Field &      Projection &             NVS & Texture Field &       Projection &             NVS &   Texture Field \\
\midrule
airplanes &     15.375 &  17.816 &   \textbf{9.236} &  \textbf{0.970} &           0.964 &         0.968 &            0.143 &           0.158 &  \textbf{0.136} \\
cars      &     70.070 &  72.209 &  \textbf{24.271} &           0.840 &  \textbf{0.887} &         0.885 &            0.236 &           0.238 &  \textbf{0.192} \\
chairs    &      8.045 &   8.788 &   \textbf{5.791} &           0.931 &  \textbf{0.947} &         0.941 &            0.127 &           0.125 &  \textbf{0.124} \\
tables    &     10.254 &   9.311 &   \textbf{8.846} &           0.934 &  \textbf{0.953} &         0.943 &            0.123 &  \textbf{0.117} &           0.123 \\
\midrule
mean      &     25.936 &  27.031 &  \textbf{12.036} &           0.919 &  \textbf{0.938} &         0.934 &            0.157 &           0.159 &  \textbf{0.144} \\
\bottomrule
\end{tabular}
	}
	\caption{
		\textbf{Single Image Texture Reconstruction.}
		Quantitative Evaluation using the FID, SSIM and Feature-$\ell_1$ metrics.}
	\label{tab:texture_reconstruction_gt_shapes}
	\vspace{-0.2cm}
\end{table*}

\begin{figure}[t!]
	\centering
	\begin{tabular}{c|c@{}c@{}c}
    \includegraphics[width=0.22\linewidth,trim={0.5cm 0.cm 0.5cm 1.5cm},clip]{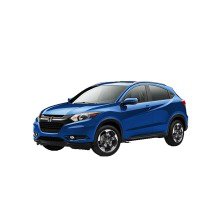} &
    \includegraphics[width=0.22\linewidth,trim={0.5cm 0.cm 0.5cm 1.5cm},clip]{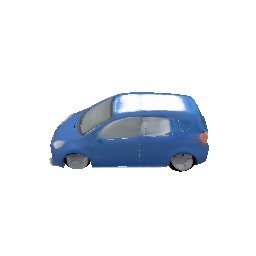} & 
    \includegraphics[width=0.22\linewidth,trim={0.5cm 0.cm 0.5cm 1.5cm},clip]{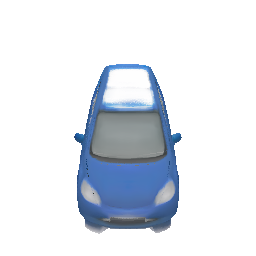} &
    \includegraphics[width=0.22\linewidth,trim={0.5cm 0.cm 0.5cm 1.5cm},clip]{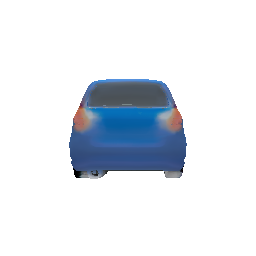} \\
   \includegraphics[width=0.22\linewidth,trim={0.5cm 0.5cm 0.5cm 1.5cm},clip]{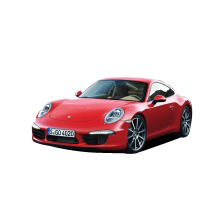} &
   \includegraphics[width=0.22\linewidth,trim={0.5cm 0.5cm 0.5cm 1.5cm},clip]{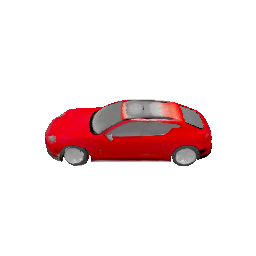} & 
   \includegraphics[width=0.22\linewidth,trim={0.5cm 0.5cm 0.5cm 1.5cm},clip]{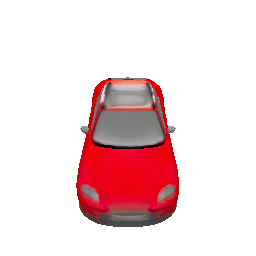} &
   \includegraphics[width=0.22\linewidth,trim={0.5cm 0.5cm 0.5cm 1.5cm},clip]{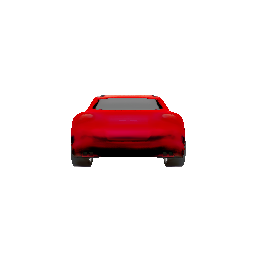} \\
    \includegraphics[width=0.22\linewidth,trim={0.5cm 0.cm 0.5cm 1.5cm},clip]{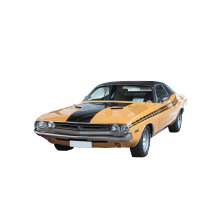} &
    \includegraphics[width=0.22\linewidth,trim={0.5cm 0.cm 0.5cm 1.5cm},clip]{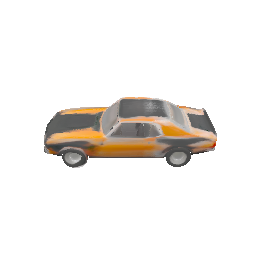} & 
    \includegraphics[width=0.22\linewidth,trim={0.5cm 0.cm 0.5cm 1.5cm},clip]{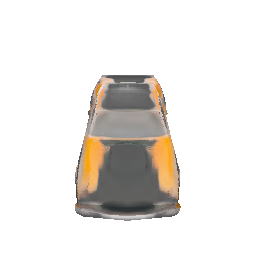} &
    \includegraphics[width=0.22\linewidth,trim={0.5cm 0.cm 0.5cm 1.5cm},clip]{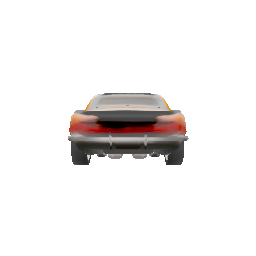} \\
   \includegraphics[width=0.22\linewidth,trim={0.5cm 0.5cm 0.5cm 1.5cm},clip]{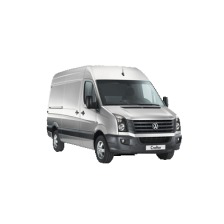} &
   \includegraphics[width=0.22\linewidth,trim={0.5cm 0.5cm 0.5cm 1.5cm},clip]{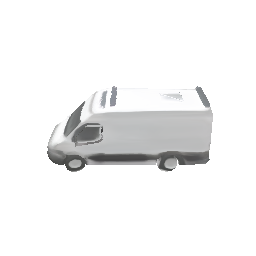} & 
   \includegraphics[width=0.22\linewidth,trim={0.5cm 0.5cm 0.5cm 1.5cm},clip]{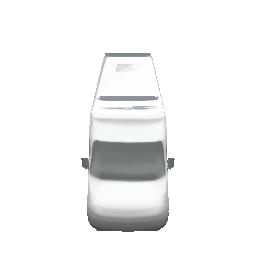} &
   \includegraphics[width=0.22\linewidth,trim={0.5cm 0.5cm 0.5cm 1.5cm},clip]{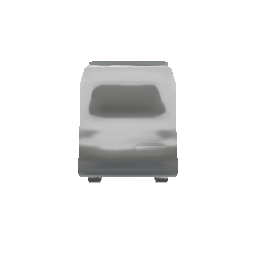} \\
    \includegraphics[width=0.22\linewidth,trim={0.5cm 0.5cm 0.5cm 0.5cm},clip]{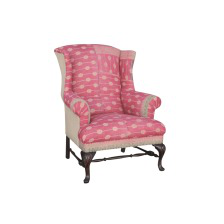} &
    \includegraphics[width=0.22\linewidth,trim={0.5cm 0.5cm 0.5cm 0.5cm},clip]{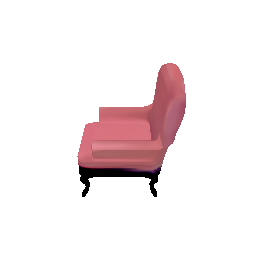} & 
    \includegraphics[width=0.22\linewidth,trim={0.5cm 0.5cm 0.5cm 0.5cm},clip]{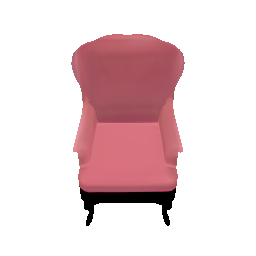} & 
    \includegraphics[width=0.22\linewidth,trim={0.5cm 0.5cm 0.5cm 0.5cm},clip]{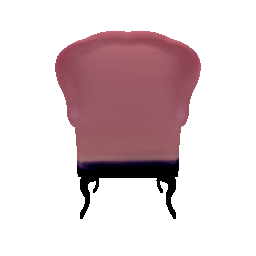} \\
    \includegraphics[width=0.22\linewidth,trim={0.5cm 0.5cm 0.5cm 0.5cm},clip]{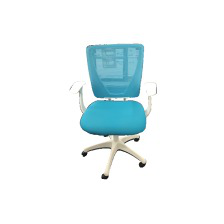} &
    \includegraphics[width=0.22\linewidth,trim={0.5cm 0.5cm 0.5cm 0.5cm},clip]{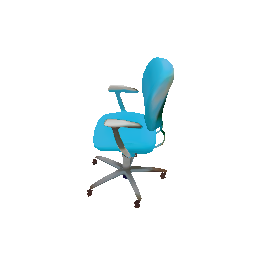} & 
    \includegraphics[width=0.22\linewidth,trim={0.5cm 0.5cm 0.5cm 0.5cm},clip]{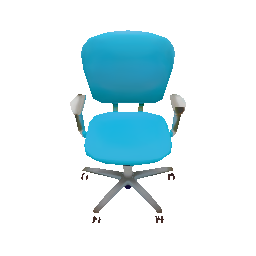} &
    \includegraphics[width=0.22\linewidth,trim={0.5cm 0.5cm 0.5cm 0.5cm},clip]{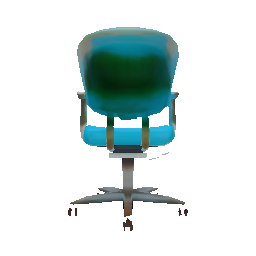} \\
    \;Condition (2D) & \multicolumn{3}{c}{Prediction (3D)}  \;
\end{tabular}

	\caption{
		\textbf{Texture Reconstruction from Real Images.}
		In this experiment, our model transfers texture from previously unseen \textit{real} images to unseen 3D CAD models.}
	\label{fig:single_image_ours_real}
	\vspace{-0.6cm}
\end{figure}

\begin{table*}[t!]
	\centering
	\resizebox{\textwidth}{!}{
		\begin{tabular}{l|cccc|cccc|cccc}
\toprule
{} & \multicolumn{4}{c}{FID} & \multicolumn{4}{c}{SSIM} & \multicolumn{4}{c}{Feature-$\ell_1$} \\
{} & Projection & Im2Avatar &      NVS &    Texture Field & Projection & Im2Avatar &             NVS &   Texture Field &       Projection & Im2Avatar &    NVS &   Texture Field \\
\midrule
airplanes &     79.146 &         - &   70.592 &  \textbf{61.760} &      0.918 &         - &  \textbf{0.921} &  \textbf{0.921} &            0.230 &         - &  0.223 &  \textbf{0.216} \\
cars      &    133.411 &   149.393 &  122.622 &  \textbf{77.439} &      0.786 &     0.760 &           0.836 &  \textbf{0.837} &            0.281 &     0.290 &  0.269 &  \textbf{0.235} \\
chairs    &     37.890 &   158.243 &   48.926 &  \textbf{36.812} &      0.817 &     0.695 &           0.841 &  \textbf{0.842} &            0.213 &     0.289 &  0.218 &  \textbf{0.207} \\
tables    &     32.693 &   115.992 &   35.086 &  \textbf{30.627} &      0.855 &     0.749 &  \textbf{0.871} &           0.869 &            0.193 &     0.265 &  0.188 &  \textbf{0.186} \\
\midrule
mean      &     70.785 &   141.209 &   69.306 &  \textbf{51.659} &      0.844 &     0.734 &  \textbf{0.867} &  \textbf{0.867} &            0.229 &     0.281 &  0.225 &  \textbf{0.211} \\
\bottomrule
\end{tabular}
	}
	\caption{
		\textbf{Full Pipeline.}
		Quantitative Evaluation using the FID, SSIM and Feature-$\ell_1$ metrics.}
	\label{tab:texture_reconstruction_onet_shapes}
	\vspace{-0.3cm}
\end{table*}

\boldparagraph{GT Shapes}
When using ground truth shapes, our model successfully learns to extract texture from just a single image of the object as illustrated in \figref{fig:single_image_gt}.
In particular, our model can complete parts of the texture that were not visible in the input view.
At the same time, our model successfully transfers texture regions visible in the input view to the shape (\eg windows and tires of the car). 
In contrast, both the projection baseline and NVS show considerable artifacts. 
While the projection baseline transfers texture incorrectly into ambiguous regions, NVS often leads to unrealistic image artifacts such as rough edges and bleeding colors.
A quantitative comparison is provided in \tabref{tab:texture_reconstruction_gt_shapes}.
While NVS achieves the best SSIM, our method performs best in terms of FID and Features-$\ell_1$ distance. 
This is consistent with the qualitative result in  \figref{fig:single_image_gt}, as SSIM is a local score whereas FID and the  Features-$\ell_1$ distance are global scores that better capture visual similarity and realism.

\begin{figure*}[t!]
	\centering
	\begin{tabular}{c@{}c@{}c@{}c@{}c@{}c@{}c}
    \includegraphics[width=0.145\linewidth,trim={1.6cm 2.5cm 0.5cm 2.5cm},clip]{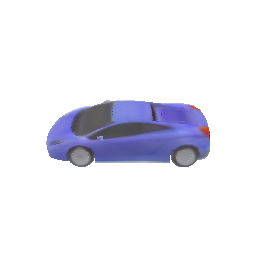} & 
    \includegraphics[width=0.17\linewidth,trim={.5cm 2.5cm 0.5cm 2.5cm},clip]{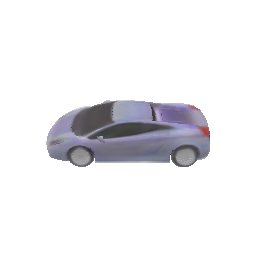} & 
    \includegraphics[width=0.17\linewidth,trim={.5cm 2.5cm 0.5cm 2.5cm},clip]{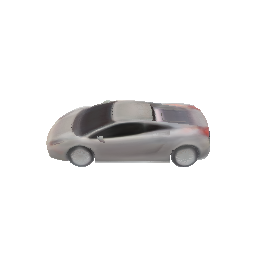} & 
    \includegraphics[width=0.17\linewidth,trim={.5cm 2.5cm 0.5cm 2.5cm},clip]{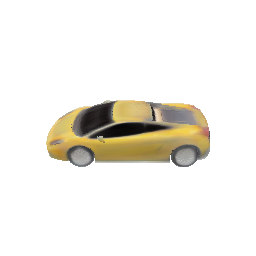} & 
    \includegraphics[width=0.17\linewidth,trim={.5cm 2.5cm 0.5cm 2.5cm},clip]{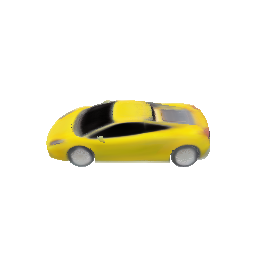}& 
    \includegraphics[width=0.145\linewidth,trim={.5cm 2.5cm 1.5cm 2.5cm},clip]{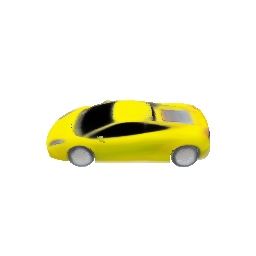}\\
    \includegraphics[width=0.145\linewidth,trim={1.5cm 2.5cm 0.5cm 2.5cm},clip]{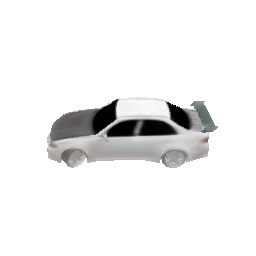} & 
    \includegraphics[width=0.17\linewidth,trim={.5cm 2.5cm 0.5cm 2.5cm},clip]{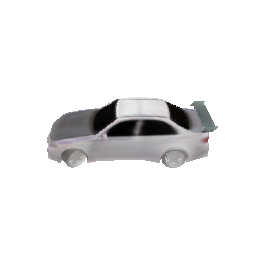} & 
    \includegraphics[width=0.17\linewidth,trim={.5cm 2.5cm 0.5cm 2.5cm},clip]{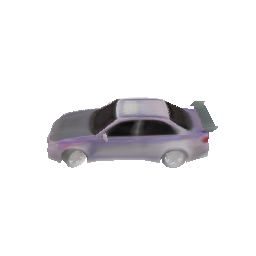} & 
    \includegraphics[width=0.17\linewidth,trim={.5cm 2.5cm 0.5cm 2.5cm},clip]{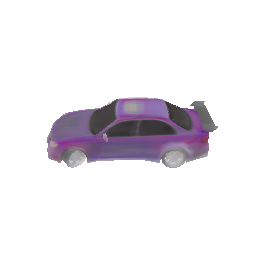} & 
    \includegraphics[width=0.17\linewidth,trim={.5cm 2.5cm 0.5cm 2.5cm},clip]{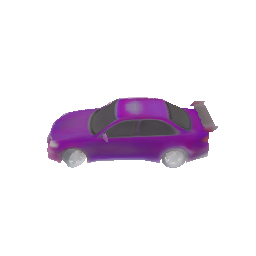}& 
    \includegraphics[width=0.145\linewidth,trim={.5cm 2.5cm 1.4cm 2.5cm},clip]{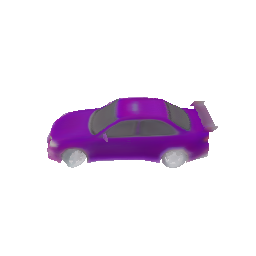}\\
    \end{tabular}

	\vspace{-0.cm}
	\caption{
		\textbf{Latent Space Interpolations (VAE).} Our generative model learns a meaningful latent space embedding.
	}
	\label{fig:vae_interpol}
	\vspace{-0.3cm}
\end{figure*}

\begin{figure}[t!]
	\centering
	\begin{tabular}{c|c@{}c@{}c}
    & Projection & NVS & Texture Field \\
   \includegraphics[width=0.22\linewidth,trim={0.5cm 1.5cm 0.5cm 2.5cm},clip]{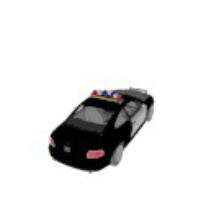} &
   \includegraphics[width=0.22\linewidth,trim={0.5cm 1.5cm 0.5cm 2.5cm},clip]{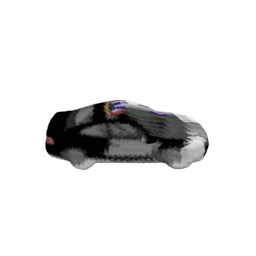} & 
   \includegraphics[width=0.22\linewidth,trim={0.5cm 1.5cm 0.5cm 2.5cm},clip]{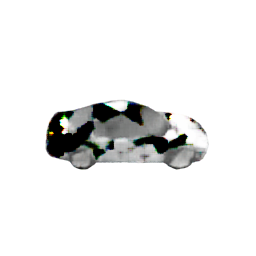} &
   \includegraphics[width=0.22\linewidth,trim={0.5cm 1.5cm 0.5cm 2.5cm},clip]{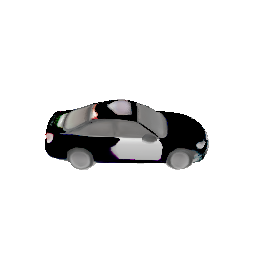} \\
   \includegraphics[width=0.22\linewidth,trim={0.5cm .cm 0.5cm 0.5cm},clip]{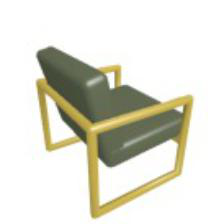} &
   \includegraphics[width=0.22\linewidth,trim={0.5cm .cm 0.5cm 0.5cm},clip]{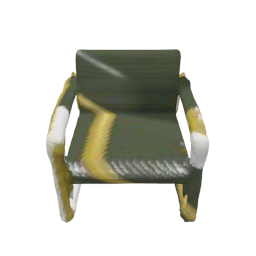} & 
   \includegraphics[width=0.22\linewidth,trim={0.5cm .cm 0.5cm 0.5cm},clip]{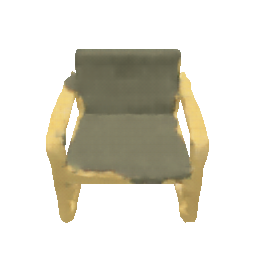} &
   \includegraphics[width=0.22\linewidth,trim={0.5cm .cm 0.5cm 0.5cm},clip]{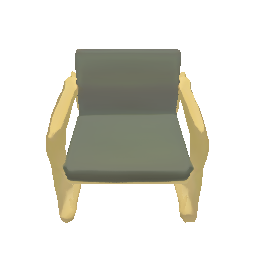} \\
   \includegraphics[width=0.22\linewidth,trim={0.7cm 1.5cm 0.5cm 2.5cm},clip]{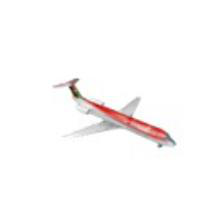} &
   \includegraphics[width=0.22\linewidth,trim={0.7cm 1.5cm 0.5cm 2.5cm},clip]{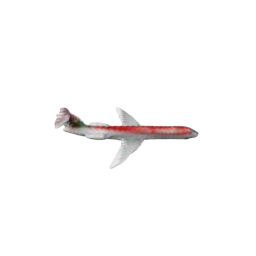} & 
   \includegraphics[width=0.22\linewidth,trim={0.7cm 1.5cm 0.5cm 2.5cm},clip]{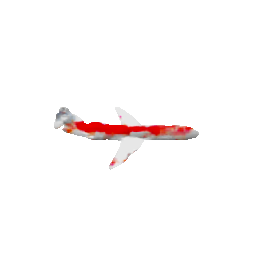} &
   \includegraphics[width=0.22\linewidth,trim={0.7cm 1.5cm 0.5cm 2.5cm},clip]{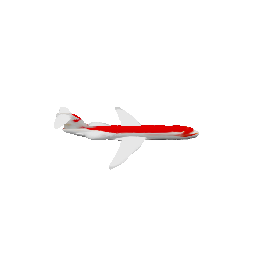} \\
   \includegraphics[width=0.22\linewidth,trim={0.5cm 0.5cm 0.5cm 1.5cm},clip]{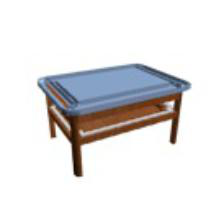} &
   \includegraphics[width=0.22\linewidth,trim={0.5cm 0.5cm 0.5cm 1.5cm},clip]{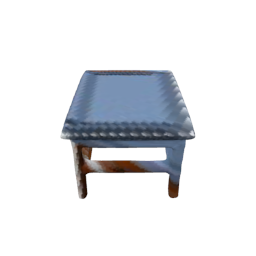} & 
   \includegraphics[width=0.22\linewidth,trim={0.5cm 0.5cm 0.5cm 1.5cm},clip]{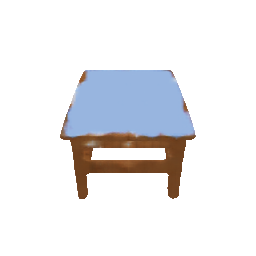} &
   \includegraphics[width=0.22\linewidth,trim={0.5cm 0.5cm 0.5cm 1.5cm},clip]{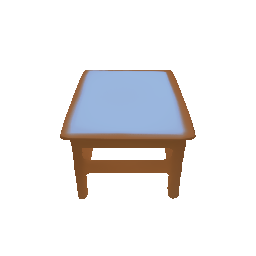} \\
    \;Condition (2D) & \multicolumn{3}{c}{Prediction (3D)}  \;
   \end{tabular}

	\caption{
		\textbf{Full Pipeline.} Results using Occupancy Networks \cite{Mescheder2019CVPR} for 3D reconstruction in combination with projection, NVS and Texture Fields for appearance estimation.
	}
	\label{fig:single_image_onet}
	\vspace{-0.2cm}
\end{figure}

\boldparagraph{Full Pipeline}
In order to obtain a full single-view textured 3D reconstruction pipeline, we combine Texture Fields with Occupancy Networks \cite{Mescheder2019CVPR}.
For a fair comparison, we also combine the projection baseline and the NVS baseline with the output from \cite{Mescheder2019CVPR}.

\begin{figure}[t!]
	\centering
	\begin{tabular}{c|cc}%{c|c@{}c}
   & Im2Avatar \cite{Sun2018ARXIV} & Texture Field \\
	\includegraphics[width=0.2\linewidth,trim={.5cm 1.5cm .5cm 1.5cm},clip]{gfx/Im2Tex/ONetShapes/car/cc7dfb5ecdf370a178c14a9d99ecf91.png}  & 
   \includegraphics[width=0.25\linewidth,trim={.5cm 2.5cm .cm 2.5cm},clip]{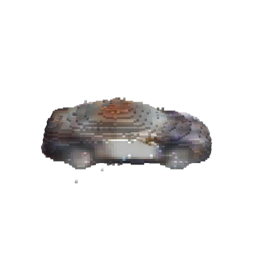} &
    \includegraphics[width=0.25\linewidth,trim={.cm 2.5cm .5cm 2.5cm},clip]{gfx/Im2Tex/ONetShapes/car/cc7dfb5ecdf370a178c14a9d99ecf91002.png} \\
    \includegraphics[width=0.2\linewidth,trim={.5cm 1.5cm .5cm 1.5cm},clip]{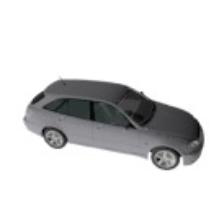} 
   & \includegraphics[width=0.25\linewidth,trim={.5cm 2.5cm .cm 2.5cm},clip]{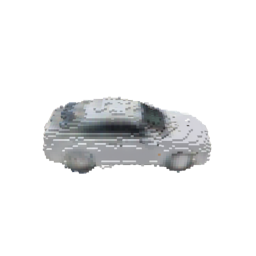} &
    \includegraphics[width=0.25\linewidth,trim={.cm 2.5cm .5cm 2.5cm},clip]{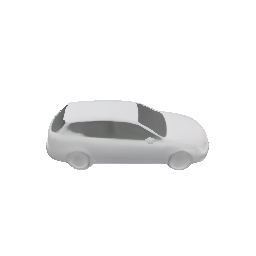} \\
    Condition (2D) & \multicolumn{2}{c}{Prediction (3D)} \\
    \end{tabular}

	\caption{
		\textbf{Voxels vs. Function Space.} We compare our full pipeline against a voxel-based reconstruction method \cite{Sun2018ARXIV}.
	}
	\label{fig:single_image_im2avatar}
	\vspace{-0.3cm}
\end{figure}

Our qualitative results in \figref{fig:single_image_onet} and \figref{fig:single_image_im2avatar} demonstrate that our approach is able to reconstruct 3D models with texture from just a single view of the model.
In contrast to Im2Avatar\footnote{%
	Unfortunately, Sun \etal provide training data only for a subset of our training data.
	We therefore train and evaluate Im2Avatar on the training and test split provided by Sun \etal.
	As the test split provided by Sun \etal and our test split are disjoint for the ``chairs'' and ``tables'' categories, we can therefore also only show a qualitative comparision on ``cars'' category. 
} \cite{Sun2018ARXIV}, NVS and the projection baseline, we observe that our method achieves more consistent and realistic looking outputs.
This is also reflected in the numerical results in \tabref{tab:texture_reconstruction_onet_shapes}:
while NVS and our method both obtain the best SSIM, our method achieves the best FID and Features-$\ell_1$ distances.

\boldparagraph{Real images}
In \figref{fig:single_image_ours_real}, we finally investigate whether our approach is also able to transfer texture information from real input images to ground truth CAD models.
Towards this goal, we apply our approach to the images provided by \cite{SunPix3D2018CVPR} and \cite{Zhu2018NIPS} and select CAD models that are similar to the object shown in the input image. 
We observe that our model generalizes reasonably well to real data despite only being trained on synthetic data.

\subsection{Unconditional Model}
In this section, we conduct unconditional experiments to investigate if Texture Fields can also be applied in a purely generative manner where we only provide the shape of the object to the network, but not the 2D image.
Towards this goal, we train both, the VAE and the GAN model, on the ``cars'' category.
During training, we provide both target images and depth maps but no input views to the network. 
During testing, we sample the latent code $z$ from a standard normal distribution to obtain random texture samples for the given 3D object.

Random samples for the GAN and the VAE model are shown in \figref{fig:gan_vae}. 
While our unconditional models successfully generate realistic textures, both the VAE and GAN samples contain artifacts similar to those present when applying VAEs and GANs to the image domain.
For example, the samples of the VAE model are globally consistent but slightly blurry.
In contrast, the samples for the GAN model are sharper but contain artifacts. 
In the future, we would like to explore combinations of VAEs and GANs and more advanced models and training methods \cite{Gregor2015ICML,Karras2018ARXIV} to improve upon these initial results.

\begin{figure}[t!]
	\centering
	\begin{tabular}{cc@{}c@{}c}
 & 
    \includegraphics[width=0.24\linewidth,trim={1.5cm 2.cm 0.5cm 2.5cm},clip]{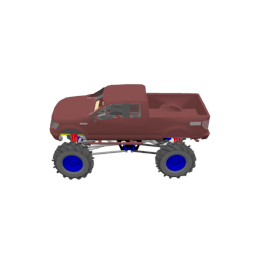} & 
    \includegraphics[width=0.24\linewidth,trim={1.5cm 2.cm 0.5cm 2.5cm},clip]{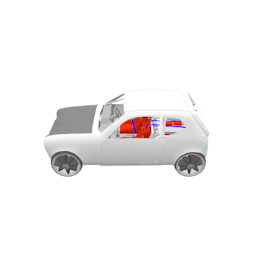} & 
    \includegraphics[width=0.22\linewidth,trim={1.5cm 2.cm 1.5cm 2.5cm},clip]{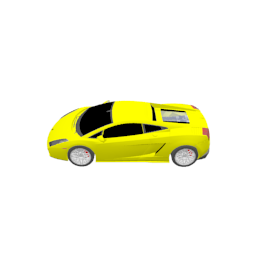} \\
    \cline{2-4}
      \multicolumn{1}{c|}{\includegraphics[width=0.22\linewidth,trim={1.5cm 2.6cm 1.5cm 2.5cm},clip]{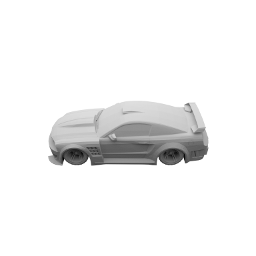} }& 
    \includegraphics[width=0.24\linewidth,trim={1.5cm 2.5cm 0.5cm 2.5cm},clip]{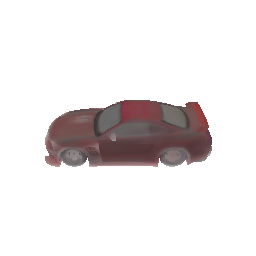} & 
    \includegraphics[width=0.24\linewidth,trim={1.5cm 2.5cm 0.5cm 2.5cm},clip]{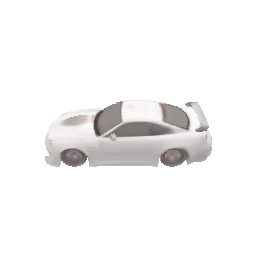} & 
    \includegraphics[width=0.21\linewidth,trim={1.5cm 2.6cm 1.5cm 2.5cm},clip]{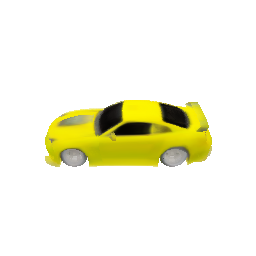} \\
       \multicolumn{1}{c|}{\includegraphics[width=0.22\linewidth,trim={1.5cm 2.6cm 1.5cm 2.5cm},clip]{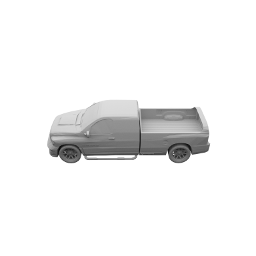} }& 
    \includegraphics[width=0.24\linewidth,trim={1.5cm 2.5cm 0.5cm 2.5cm},clip]{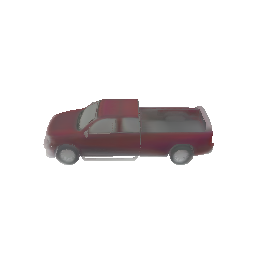} & 
    \includegraphics[width=0.24\linewidth,trim={1.5cm 2.5cm 0.5cm 2.5cm},clip]{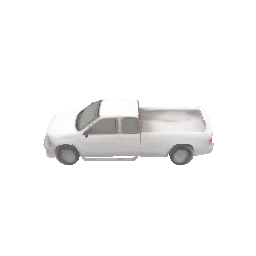} & 
    \includegraphics[width=0.21\linewidth,trim={1.5cm 2.6cm 1.5cm 2.5cm},clip]{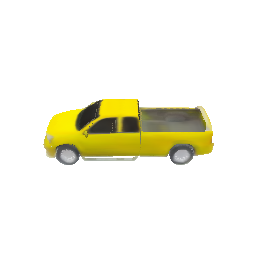} \\
    \end{tabular}

	\vspace{-0.cm}
	\caption{
		\textbf{Texture Transfer (VAE).}
		Our model transfers appearance information (top) to the target models (left).
	}
	\label{fig:vae_transfer}
	\vspace{-0.3cm}
\end{figure}

\begin{figure}[t!]
	\centering
	\begin{tabular}{c@{}c@{}c@{}c}
    \includegraphics[width=0.24\linewidth,trim={1.5cm 2.5cm 1.cm 2.5cm},clip]{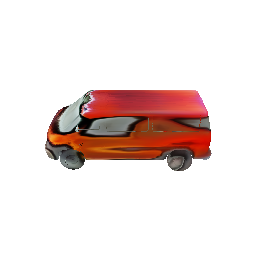} & 
    \includegraphics[width=0.24\linewidth,trim={1.5cm 2.5cm 1.cm 2.5cm},clip]{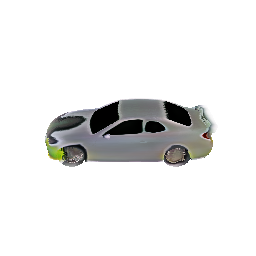} &
    \includegraphics[width=0.24\linewidth,trim={1.5cm 2.5cm 1.cm 2.5cm},clip]{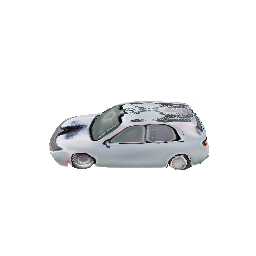} &
    \includegraphics[width=0.24\linewidth,trim={1.5cm 2.5cm 1.cm 2.5cm},clip]{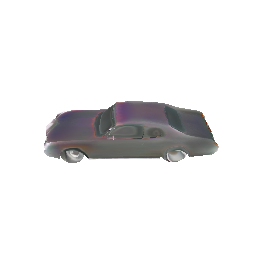} \\
    \includegraphics[width=0.24\linewidth,trim={1.5cm 2.5cm 1.cm 2.5cm},clip]{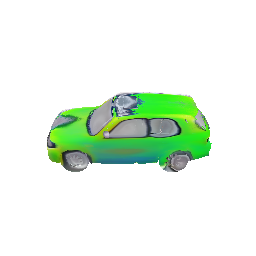} &
    \includegraphics[width=0.24\linewidth,trim={1.5cm 2.5cm 1.cm 2.5cm},clip]{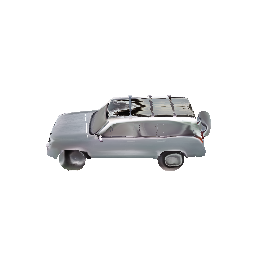} &
    \includegraphics[width=0.24\linewidth,trim={1.5cm 2.5cm 1.cm 2.5cm},clip]{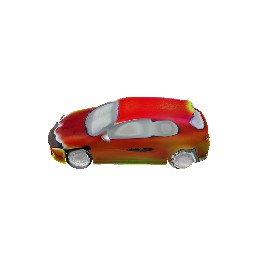} & 
    \includegraphics[width=0.24\linewidth,trim={1.5cm 2.5cm 1.cm 2.5cm},clip]{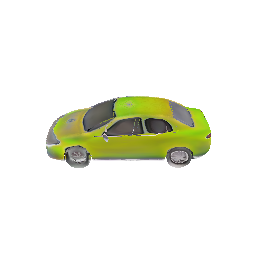}\\
\end{tabular}

	\rule{\linewidth}{0.3px}
	\begin{tabular}{c@{}c@{}c@{}c}
    \includegraphics[width=0.24\linewidth,trim={1.cm 2.5cm 1.cm 2.5cm},clip]{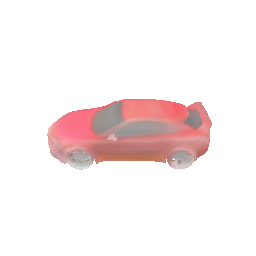} & 
    \includegraphics[width=0.24\linewidth,trim={1.cm 2.5cm 1.cm 2.5cm},clip]{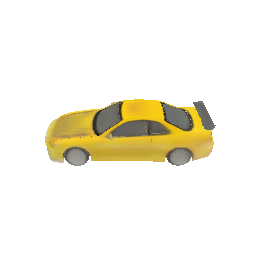} &
    \includegraphics[width=0.24\linewidth,trim={1.cm 2.5cm 1.cm 2.5cm},clip]{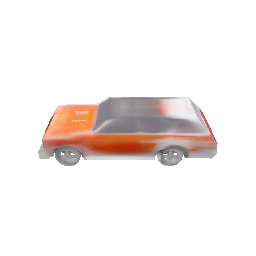} &
    \includegraphics[width=0.24\linewidth,trim={1.cm 2.5cm 1.cm 2.5cm},clip]{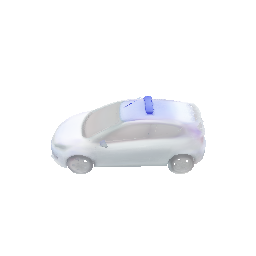}\\ 
    \includegraphics[width=0.24\linewidth,trim={1.cm 2.5cm 1.cm 2.5cm},clip]{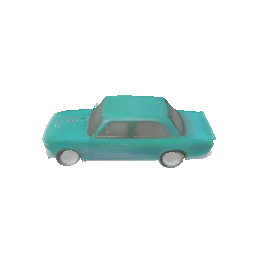} &
    \includegraphics[width=0.24\linewidth,trim={1.cm 2.5cm 1.cm 2.5cm},clip]{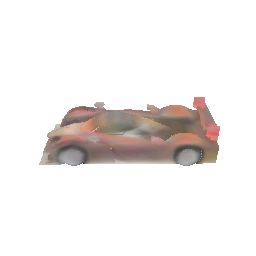}&
    \includegraphics[width=0.24\linewidth,trim={1.cm 2.5cm 1.cm 2.5cm},clip]{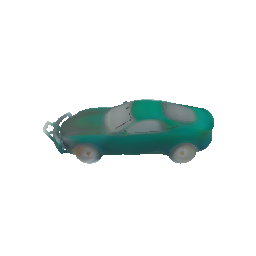} & 
    \includegraphics[width=0.24\linewidth,trim={1.cm 2.5cm 1.cm 2.5cm},clip]{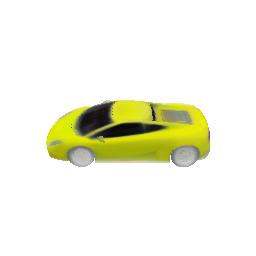}\\
    \end{tabular}
	\caption{
		\textbf{Generative Model.} Textures generated using the GAN (top 2 rows) and VAE (bottom 2 rows) models. Note that no 2D image is provided as input in this setting.}
	\label{fig:gan_vae}
	\vspace{-0.3cm}
\end{figure}

\figref{fig:vae_interpol} shows interpolations in the latent space for the VAE model.
We see that the VAE has learned a meaningful latent space and can hence interpolate smoothly between different texture samples.
Moreover, in \figref{fig:vae_transfer}, we demonstrate that our VAE model also allows for successfully transferring texture from one model to another one.

\section{Conclusion}
In this paper we introduced Texture Fields, a novel continuous representation for texture of 3D shapes.
Our experiments show that Texture Fields can predict high frequency textures from just a single object view.
Moreover, we have demonstrated that Texture Fields can also be used in an unconditional setting where we are only given the shape of the 3D object.
We hence believe that Texture Fields are a useful representation for 3D reconstruction and hope that they will become an integral part of the next generation of 3D generative models.

\section*{Acknowledgements}
This work was supported by Microsoft Research through its PhD Scholarship Program, by the Intel Network on Intelligent Systems and by an NVIDIA research gift.
\FloatBarrier

{\small
\bibliographystyle{ieee}
\bibliography{bibliography_long,bibliography,bibliography_custom}
}

\onecolumn
\appendix
\section*{\LARGE Supplementary Material}

\setcounter{section}{0}
\section{Implementation Details}
In this section, we provide more information about the network architectures used in the experiments. Furthermore, we explain the architecture of the novel view synthesis baseline and provide more information about the pipeline for image based training.

\subsection{Architectures}
In this section, we describe the architectures of each part of our model, shown in Figure 2 of the main paper.

\begin{figure}[h]
\centering
\includegraphics[scale=0.8]{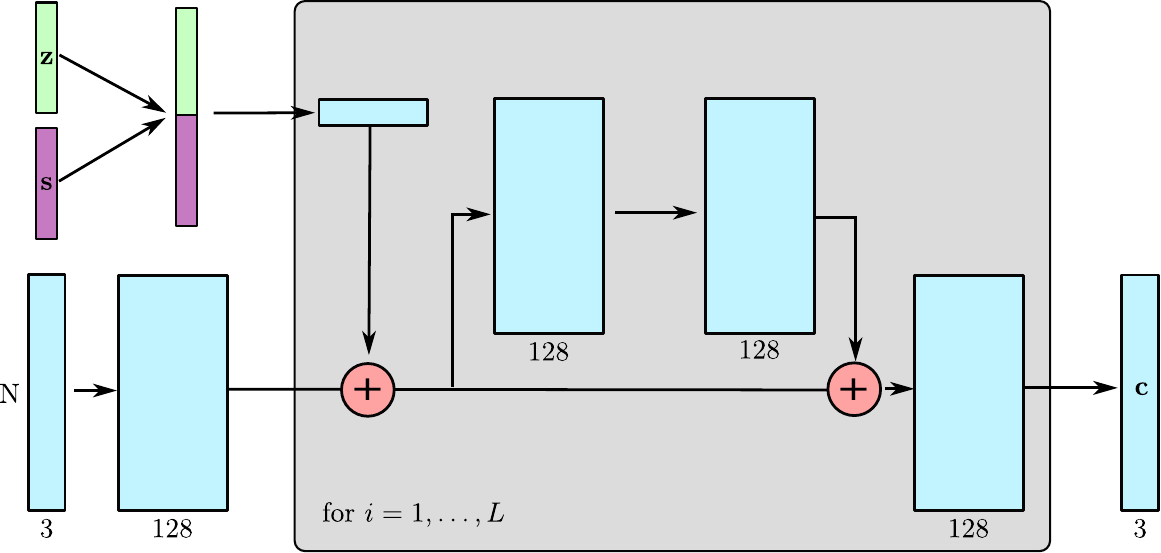}
\caption{\textbf{Texture Field.} This figure illustrates the architecture of Texture Fields. Shape Embedding $\bs$ and condition/latent texture code $\bz$ are injected to each ResNet block. For each of the $N$ 3D points, the network outputs a 3-dimensional color value $\bc$.}
\label{fig:texfield}
\end{figure}

\boldparagraph{Texture Field}
In \figref{fig:texfield}, the network architecture of the Texture Field is shown. 
We adapt the architecture proposed in \cite{Mescheder2019CVPR} for the task of texture prediction.
This architecture is used for all of our experiments.
The inputs to a Texture Field are a 3D position $\bp$, shape embedding $\bs$ and a condition or latent code $\bz$.
The shape embedding provides information about the global shape to the network, whereas $\bz$ is used for the image condition.
\figref{fig:texfield} shows the architecture applied for a set of $N$ 3D locations of a single 3D model.  
All points are passed through a fully-connected neural network that outputs a feature vector for each point.
The next parts of our architecture consists of ResNet blocks with fully-connected layers \cite{He2016CVPR}.
In each block we first inject features of $\bs$ and $\bz$ by concatenating them, passing them through a fully-connected network and adding the output to the features of each point.
We apply $L=6$ ResNet blocks for the single image texture reconstruction experiments and $L=4$ ResNet blocks for the generative models.
Finally, a fully-connected layer maps the 128-dimensional feature vector to the image space. 

\begin{figure}[h]
\centering
\includegraphics[scale=0.8]{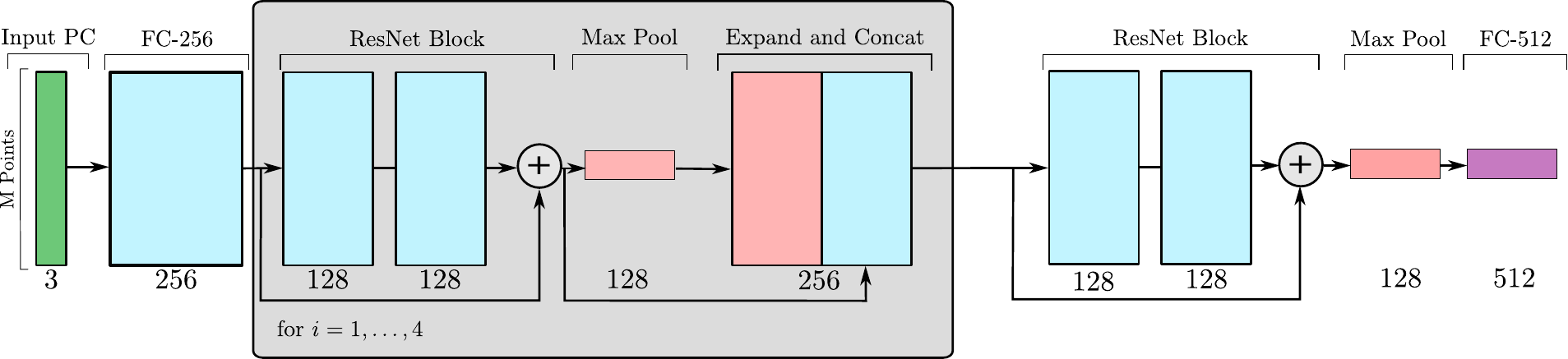}
\caption{\textbf{Shape Encoder.} Similar to a PointNet encoder \cite{Qi2017CVPR} , the network shown here determines a feature vector from a set of points. We use the ResNet-based version, proposed in \cite{Mescheder2019CVPR}.}
\label{ffig:shapeencoder}
\end{figure}

\boldparagraph{Shape Encoder}
For the pointcloud, we sample 2048 points uniformly on the surface of the 3D models.
In order to derive an embedding from the pointcloud, we utilize the pointcloud encoder, proposed in \cite{Mescheder2019CVPR}. 
The architecture is depicted in \figref{ffig:shapeencoder}. Based on PointNet \cite{Qi2017CVPR}, the network consists of 5 Resnet blocks with max pooling layers.

\begin{figure}[h]
\centering
\includegraphics[scale=0.2]{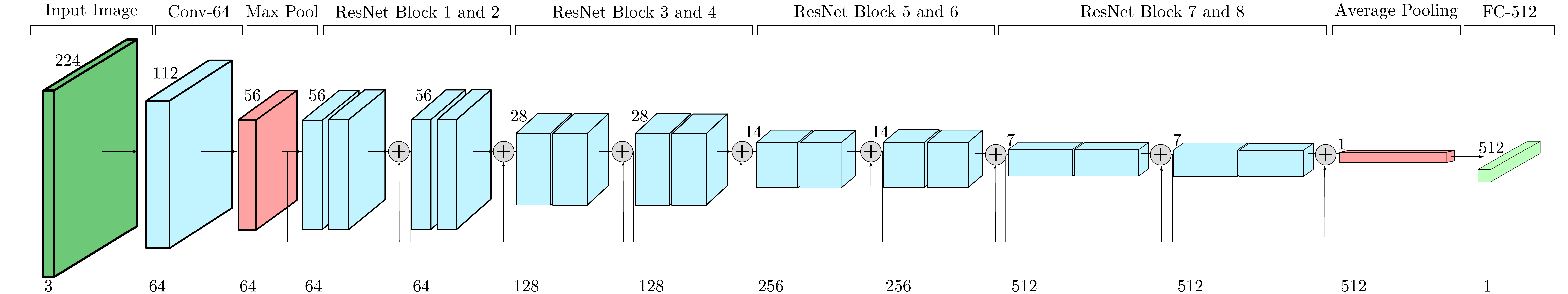}
\caption{\textbf{Image Encoder.} As encoder of the input images for the conditional task, we apply the ResNet-18 network pretrained on ImageNet.}
\label{fig:imageencoder}
\end{figure}

\boldparagraph{Image Encoder}
As encoder for the input image, we use a pretrained ResNet-18 architecture \cite{He2016CVPR}, visualized in \figref{fig:imageencoder}. After the last ResNet-Block we apply a average pooling layer and a fully-connected layer for deriving the image embedding $\bz$.

\begin{figure}[h]
\centering
\includegraphics[scale=0.19]{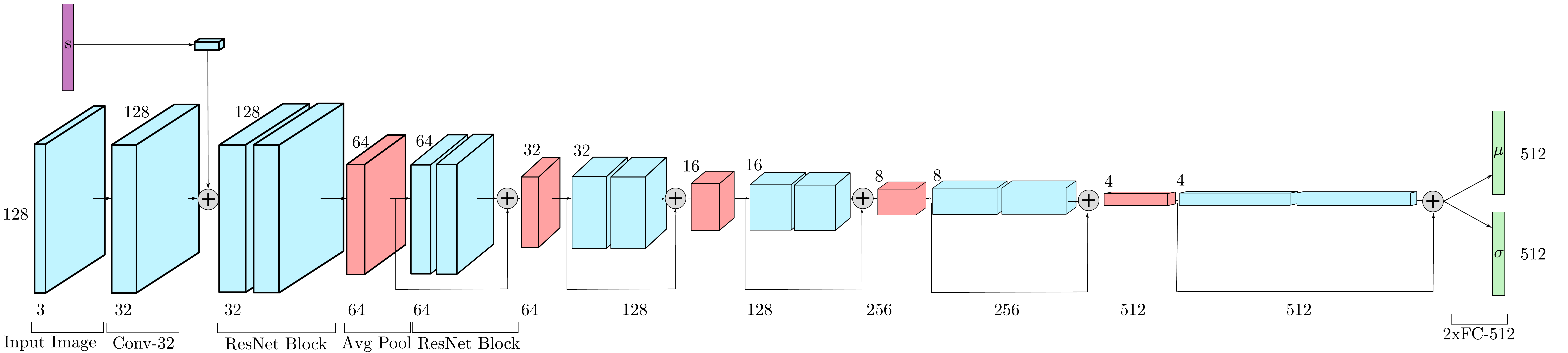}
\caption{\textbf{VAE Encoder.} Here, we illustrate the network of the encoding part of the VAE. The encoder maps an image and the shape embedding $\bs$ to mean $\mu$ and log-standard deviation $\log\sigma$ of the latent variable $\bz$.}
\label{fig:vaeencoder}
\end{figure}

\boldparagraph{VAE Encoder}
For training the VAE, we encode the ground truth views using the ResNet-based network in \figref{fig:vaeencoder}. The encoder receives an image as well as the shape embedding as input and predicts mean $\mu$ and log-standard deviation $\log\sigma$ of normal- distributed random variables in the latent space.
For injecting the shape embedding, we pass $\bs$ through a fully connected network with 32 as output dimension and add the output to each feature pixel.
Then, we iteratively apply average pooling and a ResNet block for 5 times.
Finally, we use two separate fully-connected networks to map the features to mean and log-standard deviation of the latent code $\bz$. 

\begin{figure}[t!]
\centering
\includegraphics[scale=0.2]{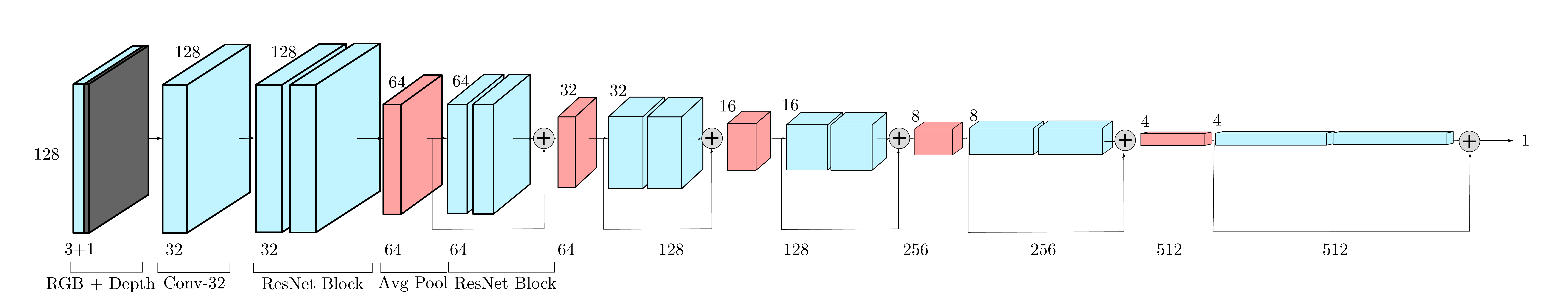}
\caption{\textbf{GAN Discriminator.} As input for the discriminator we use a RGB image and a corresponding depth image. By using average pooling and ResNet blocks, the input is mapped to a single scalar value.}
\label{fig:discriminator}
\end{figure}

\boldparagraph{GAN Discriminator}
We apply the network shown in \figref{fig:discriminator} as discriminator in the conditional GAN set up. We condition the discriminator on the depth image by concatenating the depth and RGB image. 
Using a similar architecture as in \cite{Mescheder2018ICML}, the input is mapped to a single scalar.

\subsection{NVS Baseline}
\begin{figure}[t!]
\centering
\includegraphics[scale=0.24]{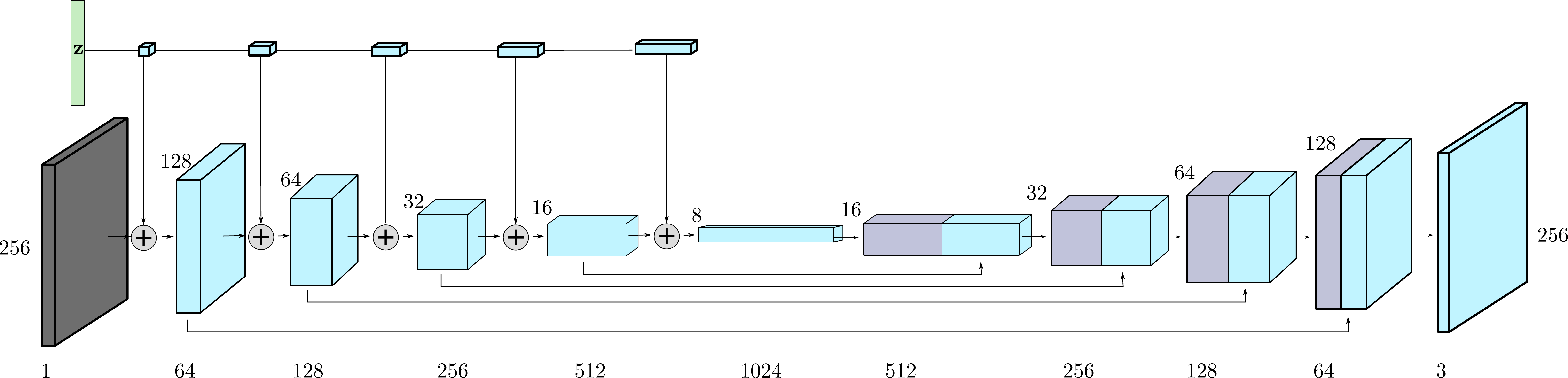}
\caption{\textbf{NVS.} In this figure, we show the U-Net-based architecture of the NVS baseline. We inject the image embedding $\bz$ into each layer in the encoding part of the network.}
\label{fig:nvs}
\vspace{-0.5cm}
\end{figure}

In \figref{fig:nvs}, we show the architecture of the novel view synthesis baseline (NVS).
The networks predict a RGB image given a depth image as input.
We apply a U-Net-based architecture \cite{Ronneberger2015MICCAI} and inject the image encoding into each layer.

\subsection{Data preparation}
We train Texture Fields from a dataset consisting of rendered images and corresponding depth images as well as intrinsic and extrinsic camera information.
To this end, we render images from 10 random views in the upper hemisphere for the 3D objects of the ShapeNet categories 'cars', 'chairs', 'airplanes' and 'tables'. 
For lighting we use a hemispheric light source.
Additionally, we render depth images from the same random views and store camera intrinsics and extrinsics, in order to be able to reproject each pixel in the depth image back to its 3D location.
In the end, the data consists of 10 views for each of the 7,499 car models, 6,781 chair models, 4,048 airplane models and 8,512 table models. For training our method we use a resolution of $128^2$. For the NVS baseline, we use $256^2$.
We evaluate every method at a resolution of $256^2$ from 10 random views. 
 
As input images for the conditional experiments, we use the renderings from Choy \etal \cite{Choy2016ECCV}.
\section{Further Results}
We present more results for each of the experiments in the following Figures \ref{fig:single_image_ours_gt_}, \ref{fig:single_image_ours_real_}, \ref{fig:single_image_ours_onet_}, \ref{fig:gan_}, \ref{fig:vae_}, \ref{fig:vae_transfer_} and \ref{fig:vae_interpol_}.

\begin{figure}[t!]
\centering
\scalebox{0.85}{
\begin{tabular}{c|c@{}c@{}c}
    \includegraphics[width=0.2\linewidth,trim={0.5cm 1.2cm 0.5cm 1.5cm},clip]{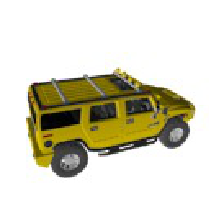} &
    \includegraphics[width=0.2\linewidth,trim={0.5cm 1.2cm 0.5cm 1.5cm},clip]{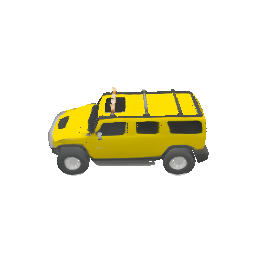} & 
    \includegraphics[width=0.2\linewidth,trim={0.5cm 1.2cm 0.5cm 1.5cm},clip]{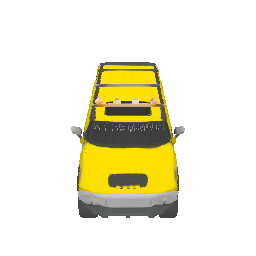} & 
    \includegraphics[width=0.2\linewidth,trim={0.5cm 1.2cm 0.5cm 1.5cm},clip]{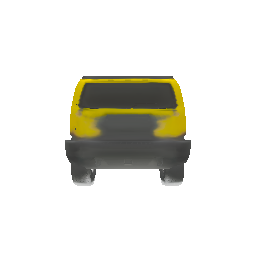} \\
        \includegraphics[width=0.2\linewidth,trim={0.5cm .5cm 0.5cm 1.5cm},clip]{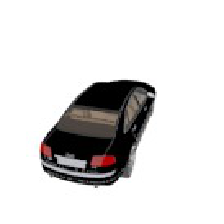} &
    \includegraphics[width=0.2\linewidth,trim={0.5cm .5cm 0.5cm 1.5cm},clip]{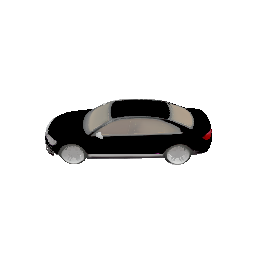} & 
    \includegraphics[width=0.2\linewidth,trim={0.5cm .5cm 0.5cm 1.5cm},clip]{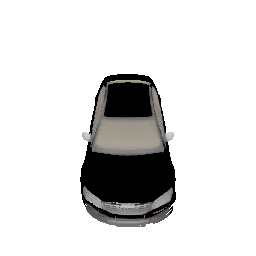} &
    \includegraphics[width=0.2\linewidth,trim={0.5cm .5cm 0.5cm 1.5cm},clip]{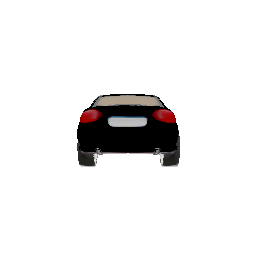} \\
    \includegraphics[width=0.2\linewidth,trim={0.5cm 1.2cm 0.5cm 0.5cm},clip]{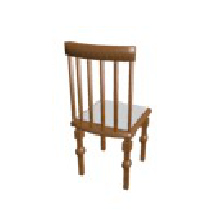} &
    \includegraphics[width=0.2\linewidth,trim={0.5cm 1.2cm 0.5cm 0.5cm},clip]{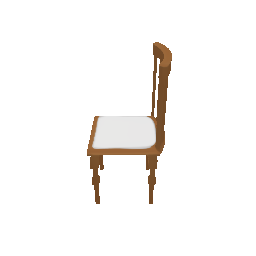} & 
    \includegraphics[width=0.2\linewidth,trim={0.5cm 1.2cm 0.5cm 0.5cm},clip]{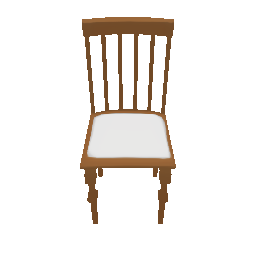} &
    \includegraphics[width=0.2\linewidth,trim={0.5cm 1.2cm 0.5cm 0.5cm},clip]{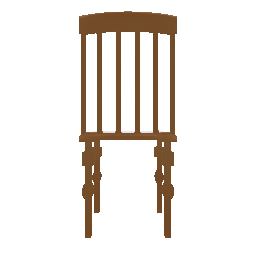} \\
    \includegraphics[width=0.2\linewidth,trim={0.5cm 1.2cm 0.5cm 0.5cm},clip]{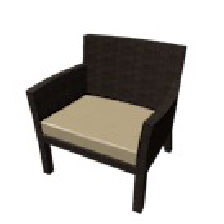} &
    \includegraphics[width=0.2\linewidth,trim={0.5cm 1.2cm 0.5cm 0.5cm},clip]{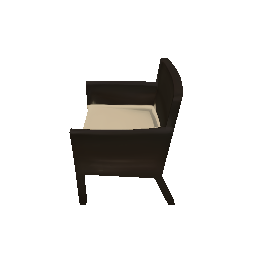} & 
    \includegraphics[width=0.2\linewidth,trim={0.5cm 1.2cm 0.5cm 0.5cm},clip]{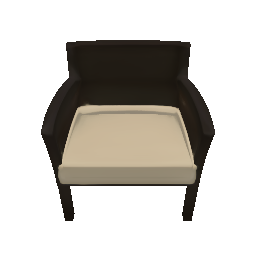} &
    \includegraphics[width=0.2\linewidth,trim={0.5cm 1.2cm 0.5cm 0.5cm},clip]{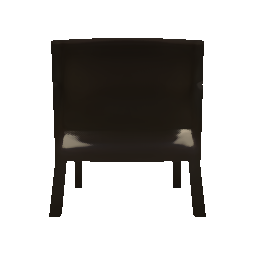} \\
    \includegraphics[width=0.2\linewidth,trim={0.5cm 1.8cm 0.5cm 1.5cm},clip]{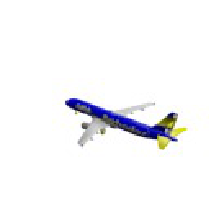} &
    \includegraphics[width=0.2\linewidth,trim={0.5cm 1.8cm 0.5cm 1.5cm},clip]{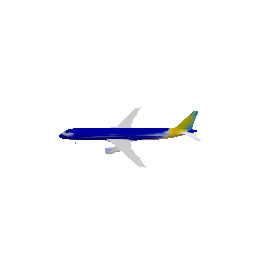} & 
    \includegraphics[width=0.2\linewidth,trim={0.5cm 1.8cm 0.5cm 1.5cm},clip]{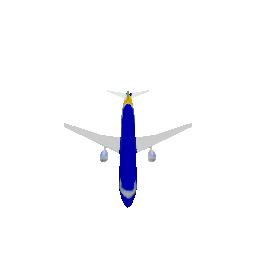} &
    \includegraphics[width=0.2\linewidth,trim={0.5cm 1.9cm 0.5cm 1.5cm},clip]{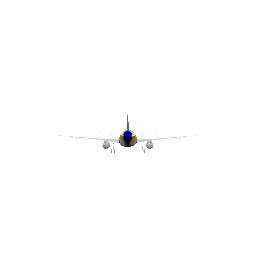} \\
    \includegraphics[width=0.2\linewidth,trim={0.5cm 1.8cm 0.5cm 1.5cm},clip]{gfx/Im2Tex/GtShapes/plane/d9dd8dd2c422dadaad70e50d5d7d02a5.png} &
    \includegraphics[width=0.2\linewidth,trim={0.5cm 1.8cm 0.5cm 1.5cm},clip]{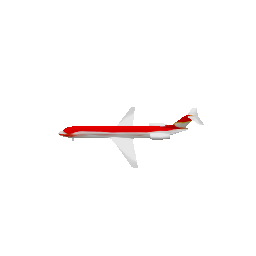} & 
    \includegraphics[width=0.2\linewidth,trim={0.5cm 1.8cm 0.5cm 1.5cm},clip]{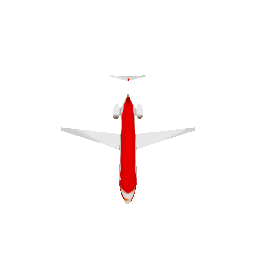} &
    \includegraphics[width=0.2\linewidth,trim={0.5cm 1.9cm 0.5cm 1.5cm},clip]{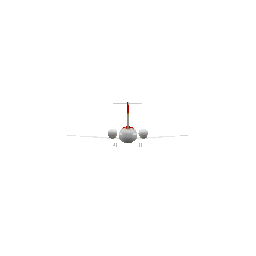} \\
    \includegraphics[width=0.2\linewidth,trim={0.5cm 1.5cm 0.5cm 1.5cm},clip]{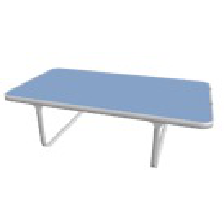} &
    \includegraphics[width=0.2\linewidth,trim={0.5cm 1.5cm 0.5cm 1.5cm},clip]{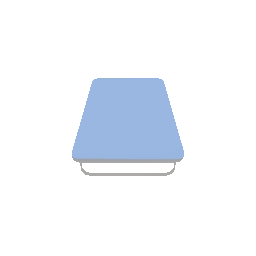} & 
    \includegraphics[width=0.2\linewidth,trim={0.2cm 1.5cm 0.2cm 1.5cm},clip]{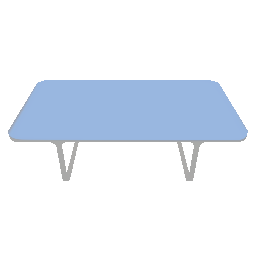} &
    \includegraphics[width=0.2\linewidth,trim={0.4cm 1.5cm 0.5cm 1.5cm},clip]{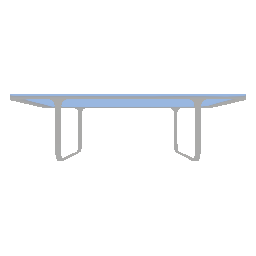} \\
        \includegraphics[width=0.2\linewidth,trim={0.5cm 0.5cm 0.5cm 1.5cm},clip]{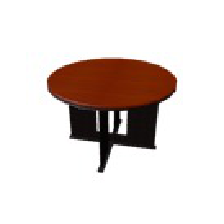} &
    \includegraphics[width=0.2\linewidth,trim={0.5cm 0.5cm 0.5cm 1.5cm},clip]{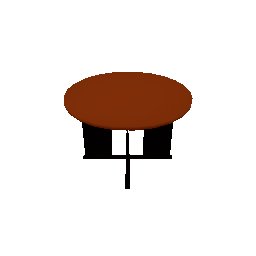} & 
    \includegraphics[width=0.2\linewidth,trim={0.2cm 0.5cm 0.2cm 1.5cm},clip]{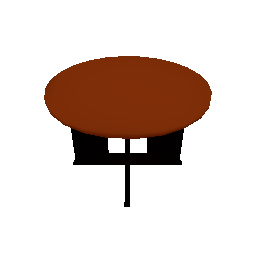} &
    \includegraphics[width=0.2\linewidth,trim={0.4cm 0.5cm 0.5cm 1.5cm},clip]{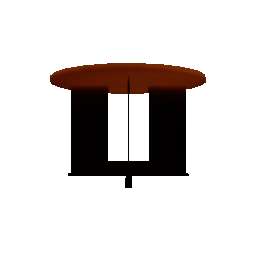} \\
    \;Condition & \multicolumn{3}{c}{Predicted}  \;
\end{tabular}}
\caption{
\textbf{Texture Reconstruction with Texture Field.} 
In this Figure, we use our model to predict texture for untextured 3D CAD models based on a single view of the same objects. 
The texture is properly predicted for all categories and contains details as lights and number plates. Very high-frequency details are sometimes leading to some blurriness.}
\label{fig:single_image_ours_gt_}
\vspace{-0.3cm}
\end{figure}

\begin{figure}[t!]
\centering
\scalebox{0.9}{
\begin{tabular}{c|c@{}c@{}c}
    \includegraphics[width=0.22\linewidth,trim={0.5cm 0.cm 0.5cm 1.5cm},clip]{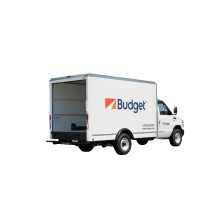} &
    \includegraphics[width=0.22\linewidth,trim={0.5cm 0.cm 0.5cm 1.5cm},clip]{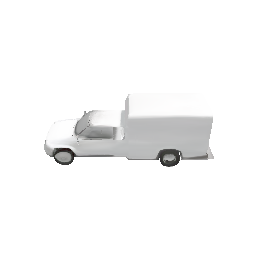} & 
    \includegraphics[width=0.22\linewidth,trim={0.5cm 0.cm 0.5cm 1.5cm},clip]{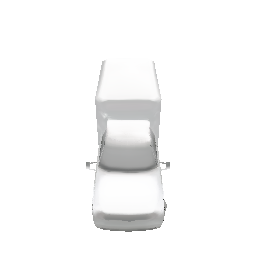} &
    \includegraphics[width=0.22\linewidth,trim={0.5cm 0.cm 0.5cm 1.5cm},clip]{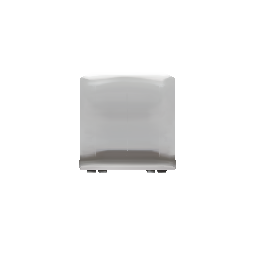} \\
   \includegraphics[width=0.22\linewidth,trim={0.5cm 0.5cm 0.5cm 1.5cm},clip]{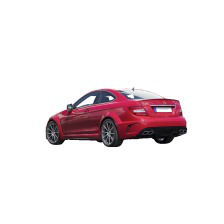} &
   \includegraphics[width=0.22\linewidth,trim={0.5cm 0.5cm 0.5cm 1.5cm},clip]{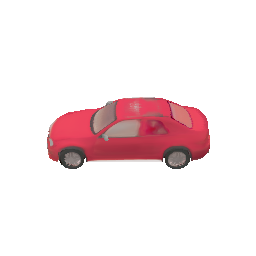} & 
   \includegraphics[width=0.22\linewidth,trim={0.5cm 0.5cm 0.5cm 1.5cm},clip]{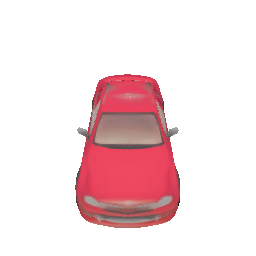} &
   \includegraphics[width=0.22\linewidth,trim={0.5cm 0.5cm 0.5cm 1.5cm},clip]{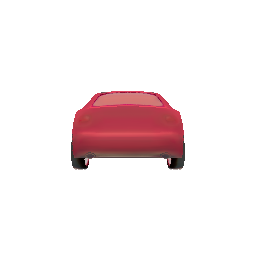} \\
    \includegraphics[width=0.22\linewidth,trim={0.5cm 0.5cm 0.5cm 0.5cm},clip]{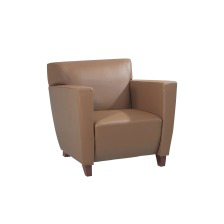} &
    \includegraphics[width=0.22\linewidth,trim={0.5cm 0.5cm 0.5cm 0.5cm},clip]{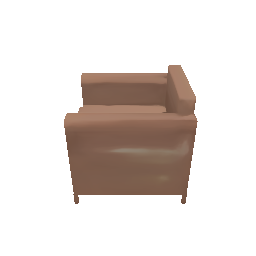} & 
    \includegraphics[width=0.22\linewidth,trim={0.5cm 0.5cm 0.5cm 0.5cm},clip]{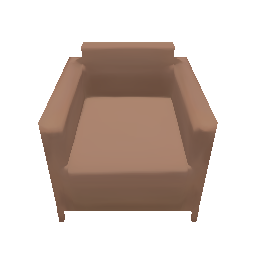} &
    \includegraphics[width=0.22\linewidth,trim={0.5cm 0.5cm 0.5cm 0.5cm},clip]{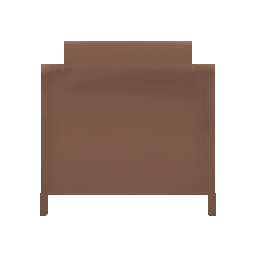} \\
    \includegraphics[width=0.22\linewidth,trim={0.5cm 0.5cm 0.5cm 0.5cm},clip]{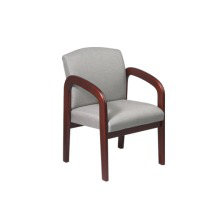} &
    \includegraphics[width=0.22\linewidth,trim={0.5cm 0.5cm 0.5cm 0.5cm},clip]{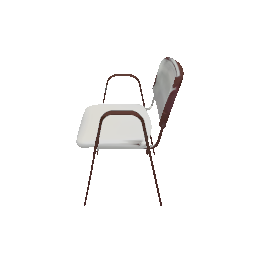} & 
    \includegraphics[width=0.22\linewidth,trim={0.5cm 0.5cm 0.5cm 0.5cm},clip]{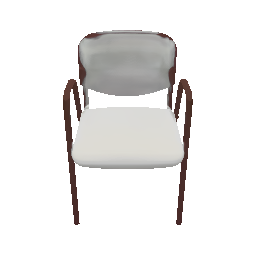} &
    \includegraphics[width=0.22\linewidth,trim={0.5cm 0.5cm 0.5cm 0.5cm},clip]{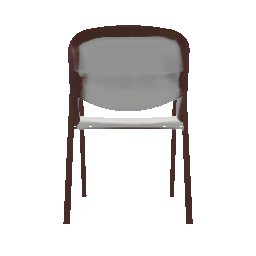} \\
    \;Condition (2D) & \multicolumn{3}{c}{Prediction (3D)}  \;
\end{tabular}}
\caption{
\textbf{Texture Reconstruction with Real Images.}
In this figure, we show results for texturing untextured synthetic CAD models from a single real input image using our approach.  
}
\label{fig:single_image_ours_real_}
\vspace{-0.3cm}
\end{figure}

\begin{figure}[t!]
\centering
\scalebox{0.9}{
\begin{tabular}{c|c@{}c@{}c}
    & Projection & NVS & Texture Field \\
   \includegraphics[width=0.22\linewidth,trim={0.5cm 1.5cm 0.5cm 2.5cm},clip]{gfx/Im2Tex/ONetShapes/car/e7c4b54fe56d9288dd1e15301c83686f.png} &
   \includegraphics[width=0.22\linewidth,trim={0.5cm 1.5cm 0.5cm 2.5cm},clip]{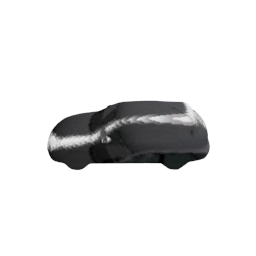} & 
   \includegraphics[width=0.22\linewidth,trim={0.5cm 1.5cm 0.5cm 2.5cm},clip]{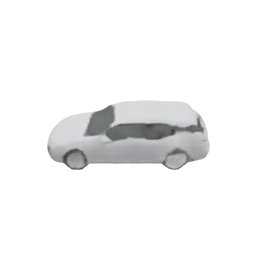} &
   \includegraphics[width=0.22\linewidth,trim={0.5cm 1.5cm 0.5cm 2.5cm},clip]{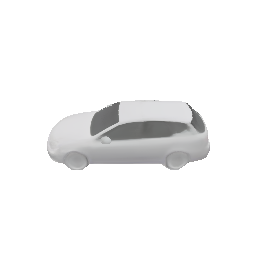} \\
      \includegraphics[width=0.22\linewidth,trim={0.5cm 1.5cm 0.5cm 2.5cm},clip]{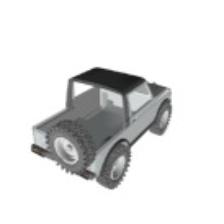} &
   \includegraphics[width=0.22\linewidth,trim={0.5cm 1.5cm 0.5cm 2.5cm},clip]{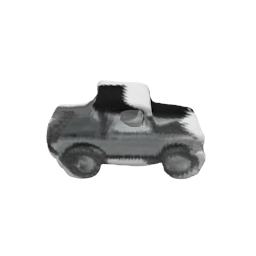} & 
   \includegraphics[width=0.22\linewidth,trim={0.5cm 1.5cm 0.5cm 2.5cm},clip]{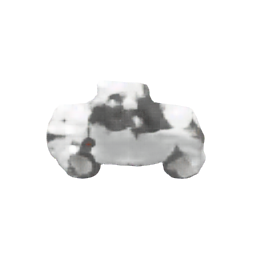} &
   \includegraphics[width=0.22\linewidth,trim={0.5cm 1.5cm 0.5cm 2.5cm},clip]{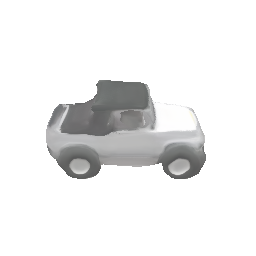} \\
      \includegraphics[width=0.22\linewidth,trim={0.5cm .5cm 0.5cm 1.5cm},clip]{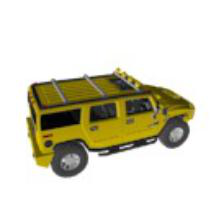} &
   \includegraphics[width=0.22\linewidth,trim={0.5cm .5cm 0.5cm 1.5cm},clip]{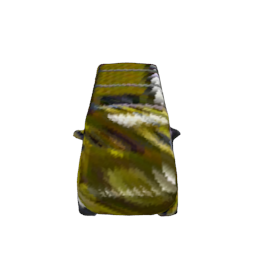} & 
   \includegraphics[width=0.22\linewidth,trim={0.5cm .5cm 0.5cm 1.5cm},clip]{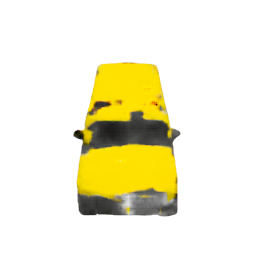} &
   \includegraphics[width=0.22\linewidth,trim={0.5cm .5cm 0.5cm 1.5cm},clip]{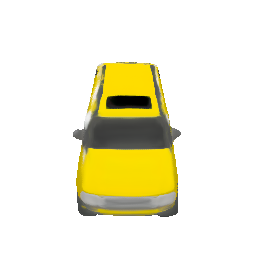} \\
      \includegraphics[width=0.22\linewidth,trim={0.5cm .5cm 0.5cm 1.5cm},clip]{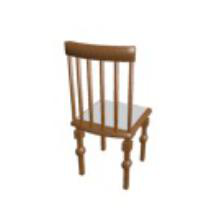} &
   \includegraphics[width=0.22\linewidth,trim={0.5cm .5cm 0.5cm 1.5cm},clip]{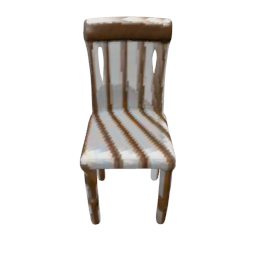} & 
   \includegraphics[width=0.22\linewidth,trim={0.5cm .5cm 0.5cm 1.5cm},clip]{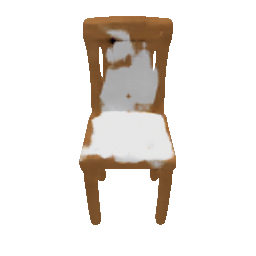} &
   \includegraphics[width=0.22\linewidth,trim={0.5cm .5cm 0.5cm 1.5cm},clip]{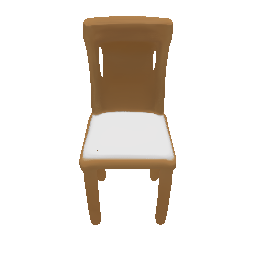} \\
      \includegraphics[width=0.22\linewidth,trim={0.5cm .5cm 0.5cm 1.5cm},clip]{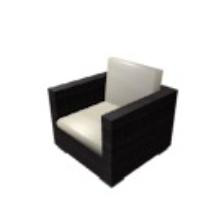} &
   \includegraphics[width=0.22\linewidth,trim={0.5cm .5cm 0.5cm 1.5cm},clip]{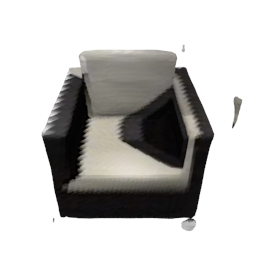} & 
   \includegraphics[width=0.22\linewidth,trim={0.5cm .5cm 0.5cm 1.5cm},clip]{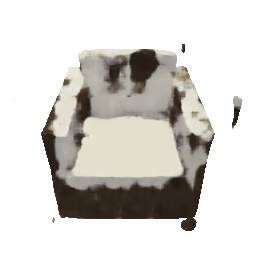} &
   \includegraphics[width=0.22\linewidth,trim={0.5cm .5cm 0.5cm 1.5cm},clip]{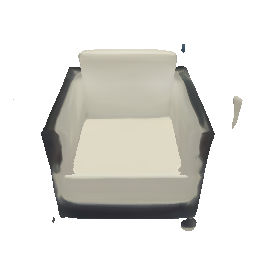} \\
      \includegraphics[width=0.22\linewidth,trim={0.5cm .5cm 0.5cm 1.5cm},clip]{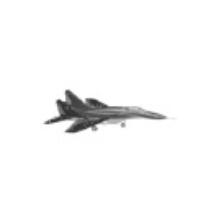} &
   \includegraphics[width=0.22\linewidth,trim={0.5cm .5cm 0.5cm 1.5cm},clip]{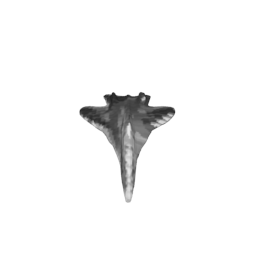} & 
   \includegraphics[width=0.22\linewidth,trim={0.5cm .5cm 0.5cm 1.5cm},clip]{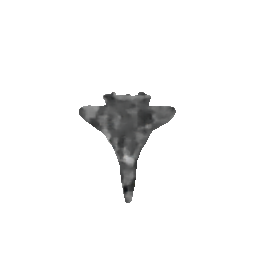} &
   \includegraphics[width=0.22\linewidth,trim={0.5cm .5cm 0.5cm 1.5cm},clip]{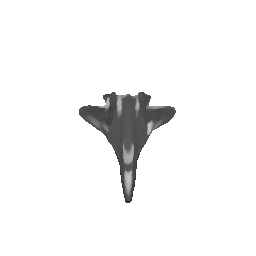} \\
         \includegraphics[width=0.22\linewidth,trim={0.5cm .5cm 0.5cm 1.5cm},clip]{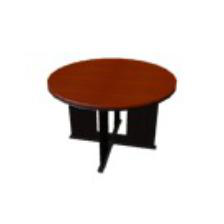} &
   \includegraphics[width=0.22\linewidth,trim={0.5cm .5cm 0.5cm 1.5cm},clip]{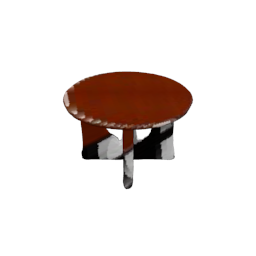} & 
   \includegraphics[width=0.22\linewidth,trim={0.5cm .5cm 0.5cm 1.5cm},clip]{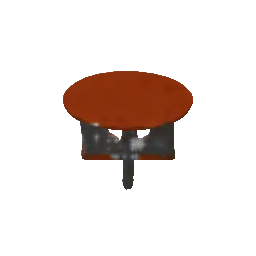} &
   \includegraphics[width=0.22\linewidth,trim={0.5cm .5cm 0.5cm 1.5cm},clip]{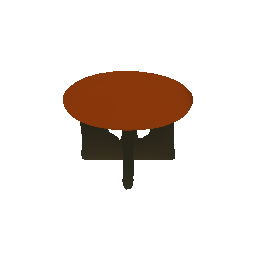} \\
    \;Condition (2D) & \multicolumn{3}{c}{Prediction (3D)}  \;
   \end{tabular}}
\caption{
\textbf{Full Pipeline.} Our full pipeline for texture and shape reconstruction leads to plausible textured 3D objects, as shown in this figure. We use the same shape reconstruction model (ONet) \cite{Mescheder2019CVPR} for all approaches.
}
\label{fig:single_image_ours_onet_}
\vspace{-0.3cm}
\end{figure}

\begin{figure}[t!]
\centering
\scalebox{0.7}{
\begin{tabular}{c@{}c@{}c@{}c}
    \includegraphics[width=0.24\linewidth,trim={1.5cm 1.cm 1.cm 2.5cm},clip]{gfx/GAN/samples/d6dab47acb946364f0cf9a4b3162f487000.png} & 
    \includegraphics[width=0.24\linewidth,trim={1.5cm 1.cm 1.cm 2.5cm},clip]{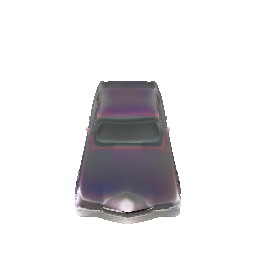} &
    \includegraphics[width=0.24\linewidth,trim={1.5cm 1.cm 1.cm 2.5cm},clip]{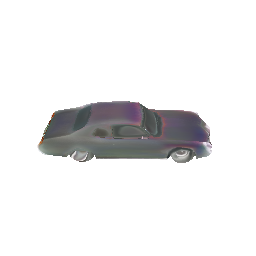} &
    \includegraphics[width=0.24\linewidth,trim={1.5cm 1.cm 1.cm 2.5cm},clip]{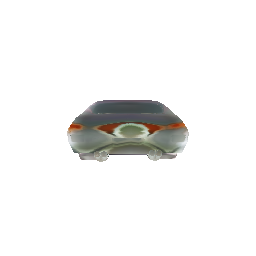} \\
    \includegraphics[width=0.24\linewidth,trim={1.5cm 1.cm 1.cm 2.5cm},clip]{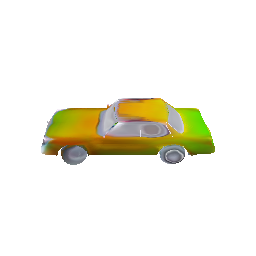} &
    \includegraphics[width=0.24\linewidth,trim={1.5cm 1.cm 1.cm 2.5cm},clip]{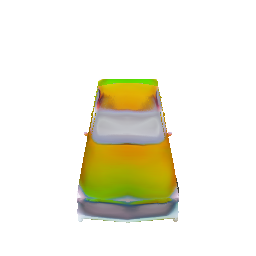} &
    \includegraphics[width=0.24\linewidth,trim={1.5cm 1.cm 1.cm 2.5cm},clip]{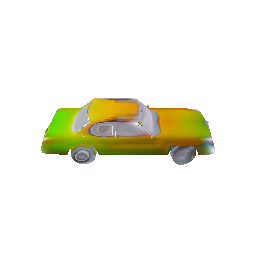} & 
    \includegraphics[width=0.24\linewidth,trim={1.5cm 1.cm 1.cm 2.5cm},clip]{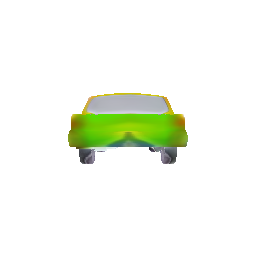}\\
        \includegraphics[width=0.24\linewidth,trim={1.5cm 1.cm 1.cm 2.5cm},clip]{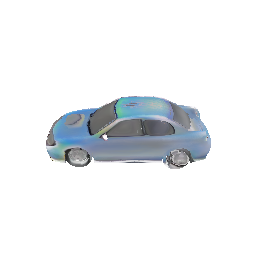} &
    \includegraphics[width=0.24\linewidth,trim={1.5cm 1.cm 1.cm 2.5cm},clip]{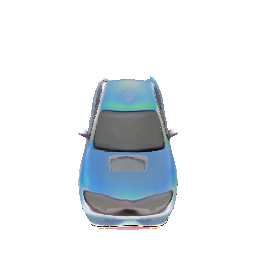} &
    \includegraphics[width=0.24\linewidth,trim={1.5cm 1.cm 1.cm 2.5cm},clip]{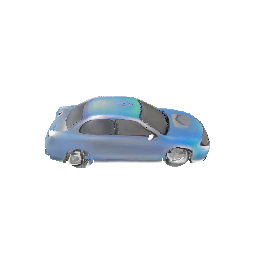} & 
    \includegraphics[width=0.24\linewidth,trim={1.5cm 1.cm 1.cm 2.5cm},clip]{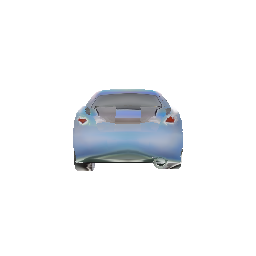}\\
        \includegraphics[width=0.24\linewidth,trim={1.5cm 1.cm 1.cm 2.5cm},clip]{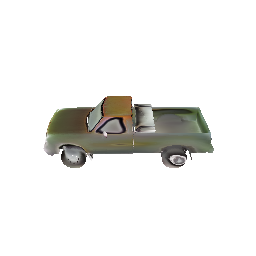} &
    \includegraphics[width=0.24\linewidth,trim={1.5cm 1.cm 1.cm 2.5cm},clip]{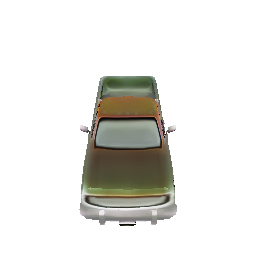} &
    \includegraphics[width=0.24\linewidth,trim={1.5cm 1.cm 1.cm 2.5cm},clip]{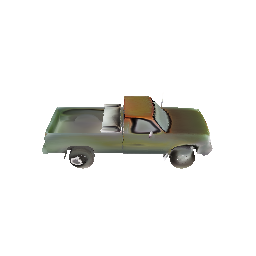} & 
    \includegraphics[width=0.24\linewidth,trim={1.5cm 1.cm 1.cm 2.5cm},clip]{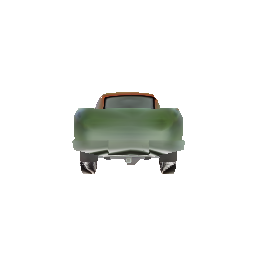}\\
\end{tabular}}
\caption{
\textbf{GAN.} In this figure, we show 3D CAD objects with generated texture using our GAN-based model. Our GAN predictions exhibits GAN typical artifacts.
}
\label{fig:gan_}
\vspace{-0.3cm}
\end{figure}

\begin{figure}[t!]
\centering
\scalebox{0.7}{
\begin{tabular}{c@{}c@{}c@{}c}
    \includegraphics[width=0.24\linewidth,trim={1.cm 1.cm 1.cm 2.cm},clip]{gfx/VAE/samples/db60ef1349d46bd62ecdadee3e6b0bc0000.png} & 
    \includegraphics[width=0.24\linewidth,trim={1.cm 1.cm 1.cm 2.cm},clip]{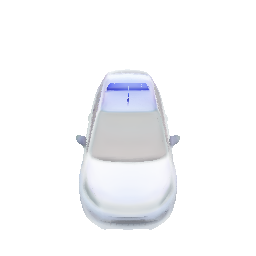} &
    \includegraphics[width=0.24\linewidth,trim={1.cm 1.cm 1.cm 2.cm},clip]{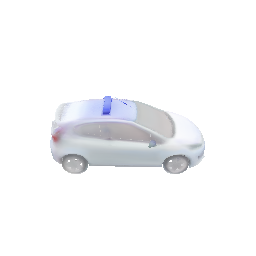} &
    \includegraphics[width=0.24\linewidth,trim={1.cm 1.cm 1.cm 2.cm},clip]{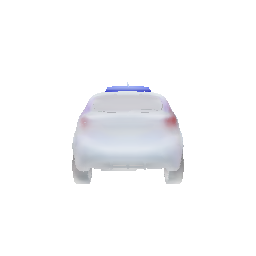}\\ 
    \includegraphics[width=0.24\linewidth,trim={1.cm 1.cm 1.cm 2.cm},clip]{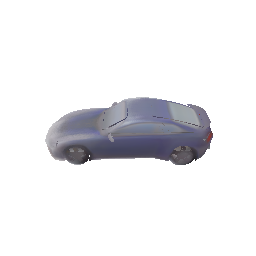} &
    \includegraphics[width=0.24\linewidth,trim={1.cm 1.cm 1.cm 2.cm},clip]{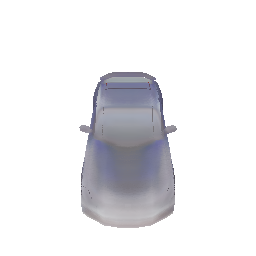}&
    \includegraphics[width=0.24\linewidth,trim={1.cm 1.cm 1.cm 2.cm},clip]{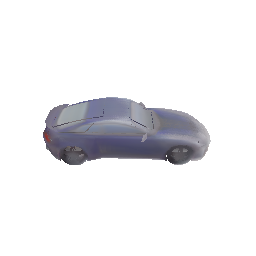} & 
    \includegraphics[width=0.24\linewidth,trim={1.cm 1.cm 1.cm 2.cm},clip]{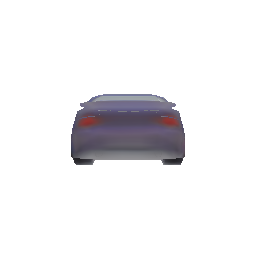}\\
        \includegraphics[width=0.24\linewidth,trim={1.cm 1.cm 1.cm 2.cm},clip]{gfx/VAE/samples/d922b4f3ba23cf43780575af49bfeda6000.png} &
    \includegraphics[width=0.24\linewidth,trim={1.cm 1.cm 1.cm 2.cm},clip]{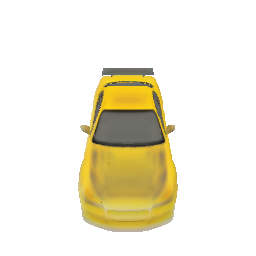}&
    \includegraphics[width=0.24\linewidth,trim={1.cm 1.cm 1.cm 2.cm},clip]{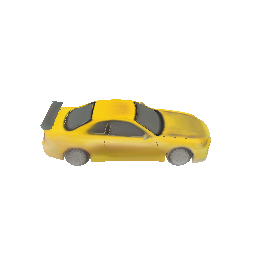} & 
    \includegraphics[width=0.24\linewidth,trim={1.cm 1.cm 1.cm 2.cm},clip]{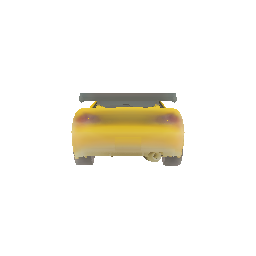}\\
            \includegraphics[width=0.24\linewidth,trim={1.cm 1.cm 1.cm 2.cm},clip]{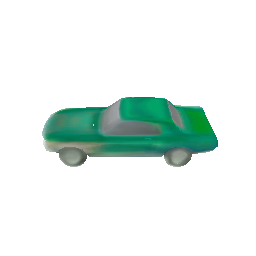} &
    \includegraphics[width=0.24\linewidth,trim={1.cm 1.cm 1.cm 2.cm},clip]{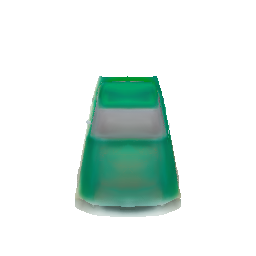}&
    \includegraphics[width=0.24\linewidth,trim={1.cm 1.cm 1.cm 2.cm},clip]{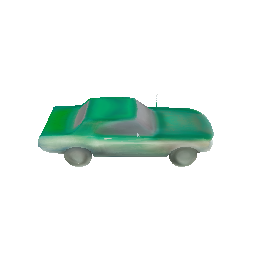} & 
    \includegraphics[width=0.24\linewidth,trim={1.cm 1.cm 1.cm 2.cm},clip]{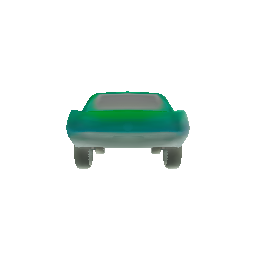}\\
    \end{tabular}}
\caption{
\textbf{VAE.} This figure illustrates predicted textures using the VAE model. The results are globally consistent, but exhibits blur in some cases.
}
\label{fig:vae_}
\vspace{-0.3cm}
\end{figure}

\begin{figure}[t!]
\centering
\begin{tabular}{c|c@{}c@{}c@{}c}
    \includegraphics[width=0.19\linewidth,trim={1.5cm 0.8cm 0.5cm 2.5cm},clip]{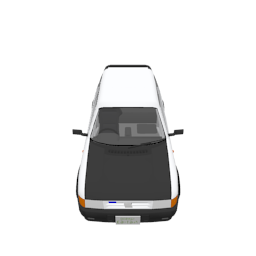} & 
    \includegraphics[width=0.19\linewidth,trim={1.5cm 0.8cm 0.5cm 2.5cm},clip]{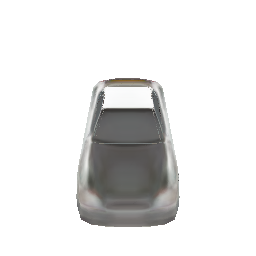} & 
    \includegraphics[width=0.19\linewidth,trim={1.5cm 0.8cm .5cm 2.5cm},clip]{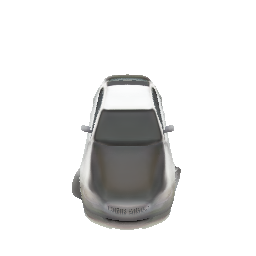} &
        \includegraphics[width=0.19\linewidth,trim={1.5cm 0.8cm 0.5cm 2.5cm},clip]{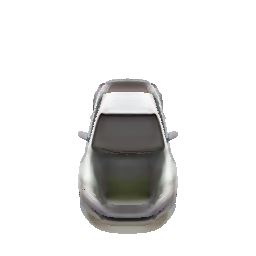} & 
    \includegraphics[width=0.19\linewidth,trim={1.5cm 0.8cm 0.5cm 2.5cm},clip]{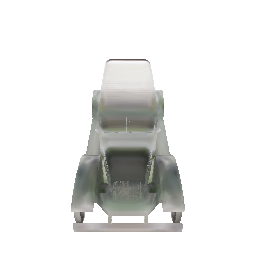} \\
        \includegraphics[width=0.19\linewidth,trim={1.5cm 0.8cm 0.5cm 2.5cm},clip]{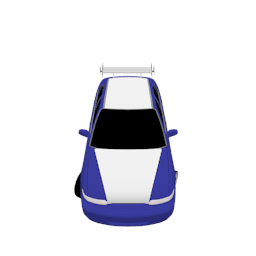} & 
    \includegraphics[width=0.19\linewidth,trim={1.5cm 0.8cm 0.5cm 2.5cm},clip]{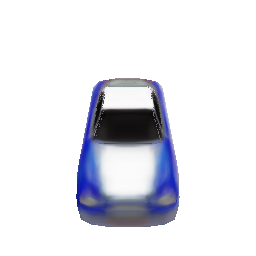} & 
    \includegraphics[width=0.19\linewidth,trim={1.5cm 0.8cm .5cm 2.5cm},clip]{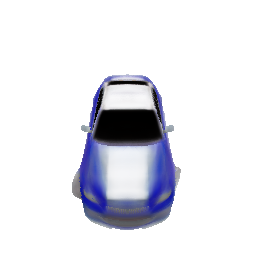} &
        \includegraphics[width=0.19\linewidth,trim={1.5cm 0.8cm 0.5cm 2.5cm},clip]{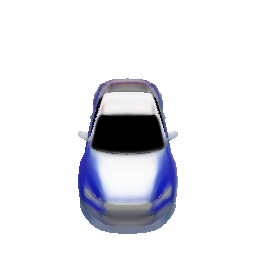} & 
    \includegraphics[width=0.19\linewidth,trim={1.5cm 0.8cm 0.5cm 2.5cm},clip]{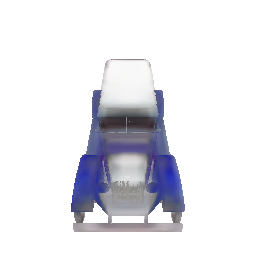} \\
        \includegraphics[width=0.19\linewidth,trim={1.5cm 0.8cm 0.5cm 2.5cm},clip]{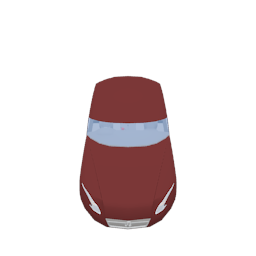} & 
    \includegraphics[width=0.19\linewidth,trim={1.5cm 0.8cm 0.5cm 2.5cm},clip]{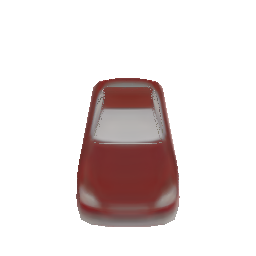} & 
    \includegraphics[width=0.19\linewidth,trim={1.5cm 0.8cm .5cm 2.5cm},clip]{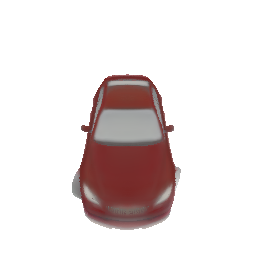} &
        \includegraphics[width=0.19\linewidth,trim={1.5cm 0.8cm 0.5cm 2.5cm},clip]{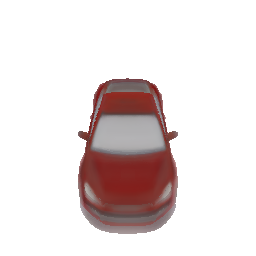} & 
    \includegraphics[width=0.19\linewidth,trim={1.5cm 0.8cm 0.5cm 2.5cm},clip]{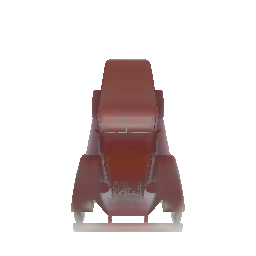} \\    
    \end{tabular}

\caption{
\textbf{Texture Transfer.} We utilize the VAE model for texture transfer from one car to another. We encode the image on the left and use the latent code for synthesizing texture for different shapes.}
\label{fig:vae_transfer_}
\vspace{-0.3cm}
\end{figure}

\begin{figure}[t!]
\centering
\begin{tabular}{c@{}c@{}c@{}c@{}c@{}c@{}c}
    \includegraphics[width=0.145\linewidth,trim={1.6cm 2.5cm 0.5cm 2.5cm},clip]{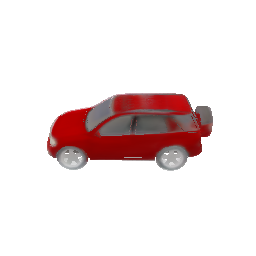} & 
    \includegraphics[width=0.17\linewidth,trim={.5cm 2.5cm 0.5cm 2.5cm},clip]{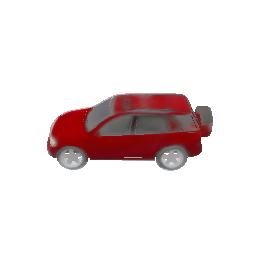} & 
    \includegraphics[width=0.17\linewidth,trim={.5cm 2.5cm 0.5cm 2.5cm},clip]{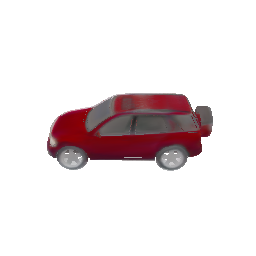} & 
    \includegraphics[width=0.17\linewidth,trim={.5cm 2.5cm 0.5cm 2.5cm},clip]{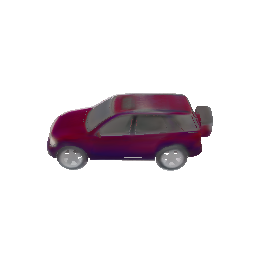} & 
    \includegraphics[width=0.17\linewidth,trim={.5cm 2.5cm 0.5cm 2.5cm},clip]{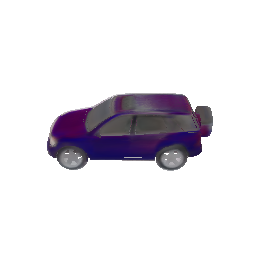}& 
    \includegraphics[width=0.145\linewidth,trim={.5cm 2.5cm 1.5cm 2.5cm},clip]{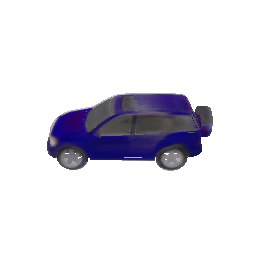}\\
    \includegraphics[width=0.145\linewidth,trim={1.5cm 2.5cm 0.5cm 2.5cm},clip]{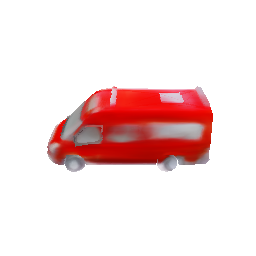} & 
    \includegraphics[width=0.17\linewidth,trim={.5cm 2.5cm 0.5cm 2.5cm},clip]{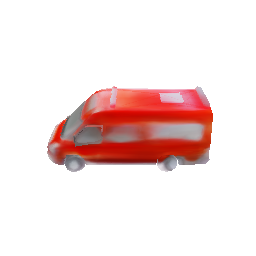} & 
    \includegraphics[width=0.17\linewidth,trim={.5cm 2.5cm 0.5cm 2.5cm},clip]{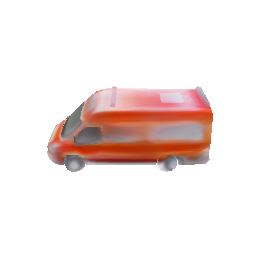} & 
    \includegraphics[width=0.17\linewidth,trim={.5cm 2.5cm 0.5cm 2.5cm},clip]{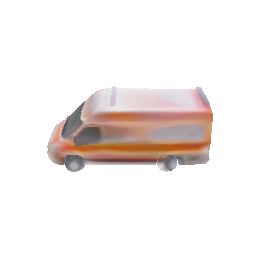} & 
    \includegraphics[width=0.17\linewidth,trim={.5cm 2.5cm 0.5cm 2.5cm},clip]{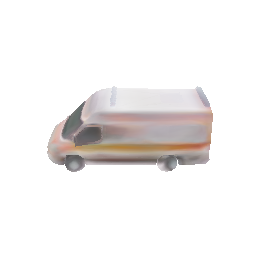}& 
    \includegraphics[width=0.145\linewidth,trim={.5cm 2.5cm 1.4cm 2.5cm},clip]{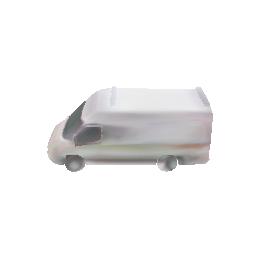}\\
    \includegraphics[width=0.145\linewidth,trim={1.5cm 2.5cm 0.5cm 2.5cm},clip]{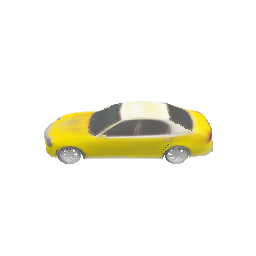} & 
    \includegraphics[width=0.17\linewidth,trim={.5cm 2.5cm 0.5cm 2.5cm},clip]{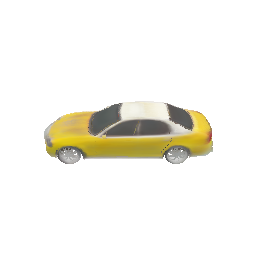} & 
    \includegraphics[width=0.17\linewidth,trim={.5cm 2.5cm 0.5cm 2.5cm},clip]{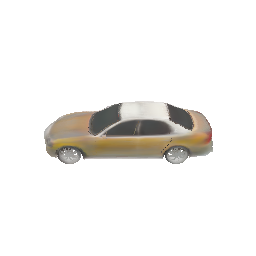} & 
    \includegraphics[width=0.17\linewidth,trim={.5cm 2.5cm 0.5cm 2.5cm},clip]{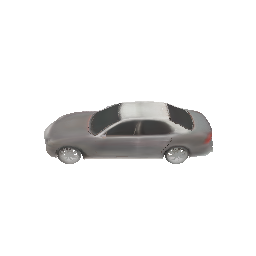} & 
    \includegraphics[width=0.17\linewidth,trim={.5cm 2.5cm 0.5cm 2.5cm},clip]{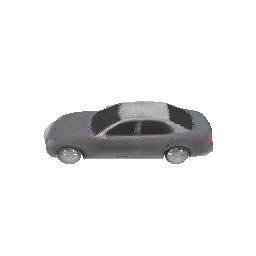}& 
    \includegraphics[width=0.145\linewidth,trim={.5cm 2.5cm 1.4cm 2.5cm},clip]{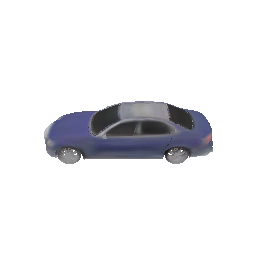}\\
    \end{tabular}

\caption{
\textbf{Latent Space Interpolations.} Here, we illustrate latent space interpolations. The results show that our model learns a continuous and meaningful latent space.
}
\label{fig:vae_interpol_}
\vspace{-0.3cm}
\end{figure}

\section{Field Visualizations}
In this section, we investigate what Texture Fields are actually learning.
For this purpose, we use the single image texture reconstruction task and we vary the input shape, while we keep the same image condition.

\begin{figure}[h]
\centering
\begin{tabular}{c|c@{}c}
{\includegraphics[scale=1]{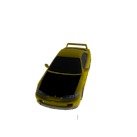}} & \includegraphics[scale=0.3]{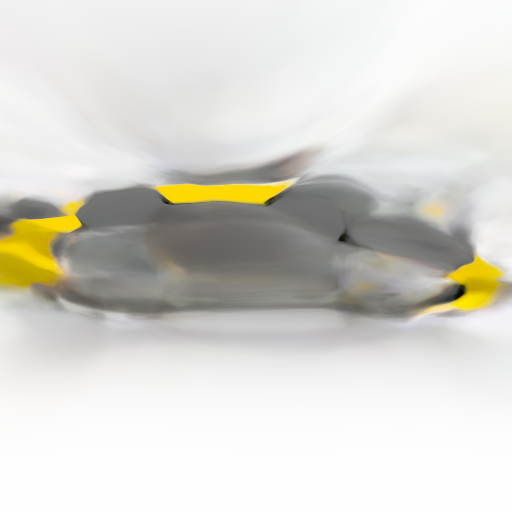} & \includegraphics[scale=0.3]{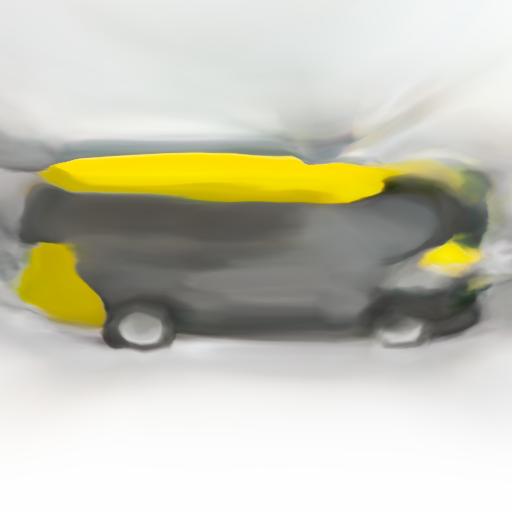}\\
 & \includegraphics[scale=0.3]{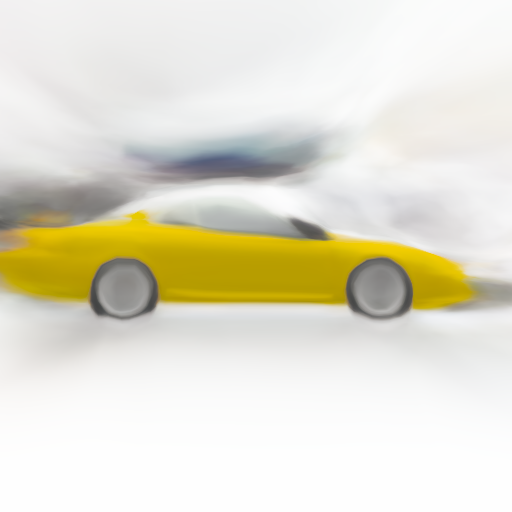} & \includegraphics[scale=0.3]{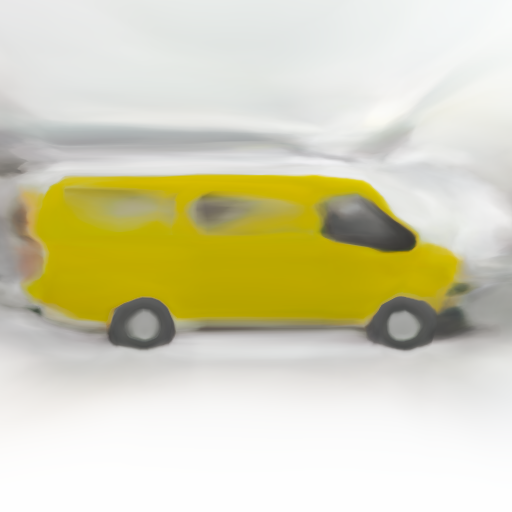}\\
 Condition & GT shape & Different shape \\
\end{tabular}
\caption{
\textbf{Texture Field Illustrations.} In this figure, we show predicted color values along cuts through the car models. The image condition is shown on the left and on the right two different results for cuts are depicted. In the top row a cut through the middle of the cars is shown, whereas in the bottom row a cut on the right side of the car. Furthermore, we show the results for two different input shapes, the corresponding 3D model and car with a completely different shape. We observe that the Texture Field learns to predict color values at locations close to the input 3D model. Far from the shape, the color values are meaningless.
}
\label{fig:tex_field_illustartions}
\vspace{-0.3cm}
\end{figure}

In \figref{fig:tex_field_illustartions}, we depict color values predicted by a Texture Field along cuts through car models. 
We see that the Texture Field learns to predict color values at the location of the shape.
As expected, color predictions far from the shape are meaningless as non of the observations constrain these areas.
In the interior of the car, a gray color usually appears, whereas outside of the car it is white.
By varying the input shape, we observe that the network is changing the locations of the color according to the shape.
The Texture Field successfully transfers texture information from the image condition onto arbitrary shapes.
This leads us to the conclusion that Texture Fields implicitly decode the shape embedding and reconstruct the texture at encoded shape locations following the image condition.
 
\end{document}